%% file: main.tex
\documentclass{article}

\usepackage[preprint]{corl_2026} %

\usepackage{times}
\usepackage{xcolor}
\usepackage{amsmath}
\usepackage{amssymb}
\usepackage{booktabs}
\usepackage{multirow}
\usepackage{graphicx}
\usepackage{mwe}
\usepackage[dvipsnames]{xcolor}
\usepackage{diagbox} %
\usepackage{enumitem}
\usepackage{algorithm}
\usepackage{algpseudocode}
\usepackage{wrapfig}

\usepackage[font=small]{caption}
\usepackage[numbers]{natbib}
\usepackage{multicol}

\usepackage{xspace}
\newcommand{\ourmethod}{\texttt{MetaGenSim}\xspace}

\newcommand{\uniformgen}{{\texttt{UniformGen}}\xspace}
\newcommand{\oracle}{{\texttt{Oracle}}\xspace}
\newcommand{\STRAP}{{\texttt{STRAP}}\xspace}
\newcommand{\IWR}{{\texttt{IWR}}\xspace}
\newcommand{\activegen}{\texttt{MetaGenSim}\xspace}

\title{Active Generative Simulation: Meta-Learning \emph{What} to Simulate for Sim-and-Real Policy Co-Training}

\author{
  Liqian Ma \quad Vaibhav Saxena \quad Utkarsh Aashu Mishra\\
  \textbf{Shuo Cheng \quad Yongxin Chen$^{*}$ \quad Danfei Xu$^{*}$}\\
  Georgia Institute of Technology\\
  {$^{*}$Equal advising}\\
}

\begin{document}
\maketitle

\vspace{-0.5cm}
\begin{abstract}
Sim-and-real co-training is an effective way to scale robot learning by augmenting limited real-world demonstrations with simulation data. However, existing approaches typically treat simulators as fixed data sources, relying on uniform generation, offline retrieval, or manual parameter alignment by domain experts. We formulate \textit{Active Generative Simulation}: optimizing a simulator's data-generation distribution to maximize downstream real-world performance. To solve this problem, we propose \ourmethod, a meta-learning framework that trains an auxiliary rater network via meta-gradients to automatically discover high-utility data-generation distributions within a user-specified simulation parameter space, without manual parameter alignment. Across simulation and real-world manipulation tasks, \ourmethod outperforms uniform generation and retrieval baselines, achieves a $150\%$ relative improvement in real-world full-task success over uniform generation, and matches expert manual alignment. We further show that distributions curated with a lightweight proxy policy transfer to VLA fine-tuning, and that \ourmethod can exploit partial alignment through atomic skill stitching and object-receptacle composition when exact sim-and-real task alignment is unavailable.
Project website: \url{https://meta-gensim.github.io/}

\end{abstract}


\input{figures/teaser/figure}

\input{sections/1_introduction}
\input{sections/2_related_works}
\input{sections/3_preliminary}

\input{sections/4_method}
\input{sections/5_experiments}

\section{Conclusion} 
\label{sec:conclusion}

We introduced \ourmethod, a meta-learning framework for Active Generative Simulation that optimizes what to simulate for sim-and-real co-training. \ourmethod learns a target-aware rater from meta-gradients, converts rater scores into data-generation distributions, and synthesizes targeted simulation datasets for improved sim-and-real co-training. Across simulation and real-world manipulation tasks, \ourmethod improves over real-only training, uniform generation, and offline retrieval, while matching the performance of expert-aligned simulation data. We further show that proxy-curated datasets transfer to VLA fine-tuning, and that \ourmethod can exploit partially aligned simulation data through atomic skill stitching and object-receptacle composition when exact sim-and-real alignment is unavailable.

\section{Limitations} 
\label{sec:limitations}
\textbf{User-specified parameter space and generation pipeline.} \ourmethod optimizes the data-generation distribution within a user-specified parameter space $\Phi$, including its ranges and the underlying generation pipeline (e.g., MimicGen source demonstrations). It can therefore only exploit factors exposed in $\Phi$ and data the pipeline can produce. Unexposed factors, such as background and lighting in our real-world setup, remain unaligned. This parameterization is straightforward in our domains but may not scale to broader simulators that represent tasks, visuals, and configurations differently. Developing a unified parameter space, potentially through learned latent parameters that make the generation pipeline more differentiable, is an exciting future direction.

\textbf{Probing coverage and density estimation.} The rater can only score regions covered by the probing dataset, which we sample uniformly in a single round. Sparse probing may miss small high-utility regions, and weighted KDE becomes less reliable as the dimension of $\Phi$ grows. Adaptive or iterative probing that refines the distribution over multiple rounds is a natural extension.

\textbf{Distribution concentration and hyperparameters.} We impose no explicit diversity or coverage constraint. Instead, the spread of the generated distribution is controlled by the temperatures and the KDE bandwidth. Overly small bandwidths or overly high temperatures degrade performance (Appendix Table~\ref{tab:ablation_stackcube}), and a narrow target dataset can yield a narrow generated distribution.

\acknowledgments{
This project is funded by NSF Award 2409016.
}

\bibliography{references}

\appendix
\include{sections/6_appendix}

\end{document}

%% file: figures/teaser/figure.tex
\begin{center}
  \includegraphics[width=0.95\linewidth]{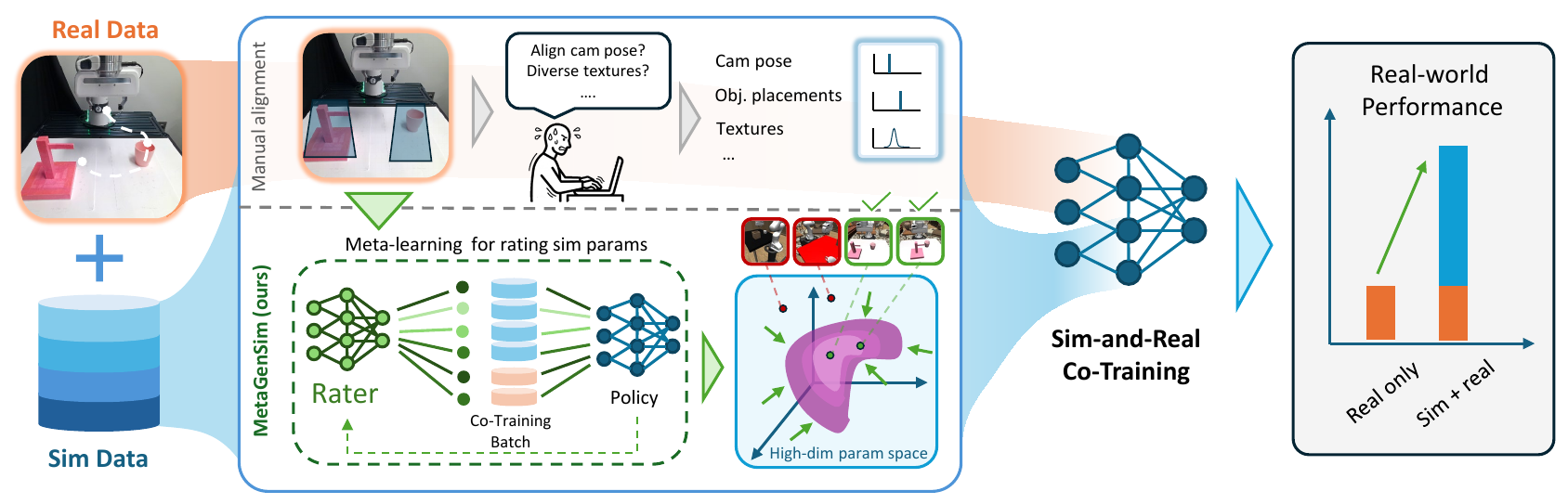}
  \captionof{figure}{
  \textbf{Active Generative Simulation via \ourmethod.} Sim-and-real co-training (top) boosts robot performance but traditionally relies on manual alignment (middle), where experts must laboriously tune data generation parameters (such as tasks, object placements, camera poses). We propose \ourmethod (bottom), which meta-learns a Rater to score synthetic data based on policy utility. This induces a utility landscape over parameters, enabling active generation of targeted simulation data that matches the real-world distribution without manually tuning the generation parameters.
  }
  \label{fig:teaser}
\end{center}

%% file: sections/1_introduction.tex
\section{Introduction}

While recent large-scale real-robot datasets have advanced generalist policies~\cite{ebert2021bridge, rt12022arxiv, black2026pi0visionlanguageactionflowmodel}, relying exclusively on real data remains expensive, labor-intensive, and safety-constrained. Generative simulation offers a scalable alternative data source by procedurally synthesizing scene layouts, assets, and tasks~\cite{Deitke2022ProcTHORLa, liang2024eurekaverse, deshpande2025graspmolmo, katara2024gen2sim, robocasa2024}. We refer to the variables controlling these environments, such as camera poses, object arrangements, visual appearance, and task goals, as \textit{data generation parameters}.

Given data generation parameters, a generative simulator can instantiate environments and generate large numbers of demonstrations with minimal human intervention~\cite{mandlekar2023mimicgen, jiang2024dexmimicgen, garrett2024skillgen, cheng2023nod, lin2026humanoidmimicgendatagenerationlocomanipulation}. However, the utility of simulation data remains limited by the sim-to-real gap. Domain randomization and system identification can mitigate this mismatch, but often require substantial human effort and cannot eliminate residual gaps~\cite{jain2025polarisscalablerealtosimevaluations}. A promising alternative is \emph{sim-and-real co-training}, where real demonstrations anchor learning to the deployment domain while simulation provides scale and diversity~\cite{ankile2024imitationrefinementresidual, wei2025simandrealcotraining, maddukuri2025simandreal, cheng2025generalizable, lei2026mechanisticanalysis}. Co-training can outperform training on real data alone~\cite{maddukuri2025simandreal, wei2025simandrealcotraining, anonymous2025robocasa}, but not all generated data is useful: misaligned camera poses, object layouts, visual factors, or task goals can provide little benefit, and exact alignment may be impossible when the simulator lacks the target setting. This raises a central question:
\textit{Among the vast space of possible simulation environments, which data should be generated to maximize the performance of a sim-and-real co-trained policy?}

Current simulation data-generation strategies largely rely on manual heuristics~\cite{mu2024adademo, gao2024efficient} or uniform randomization~\cite{torne2024reconciling,tobin2017domain}. Overly broad training distributions can degrade policy learning when many samples are irrelevant to the target task~\cite{dass2025datamil,agia2025cupid, kuhar2023learning,chen2025curating,memmel2024strap}. Prior work suggests that task-aware alignment helps~\cite{wei2025simandrealcotraining}, but manually tuning high-dimensional generation parameters is laborious, requires expert knowledge, and can be misleading when exact alignment is impossible.

In this work, we formulate \textit{Active Generative Simulation}: determining a distribution over simulation parameters such that co-training on the resulting synthetic data maximizes target-task performance. To efficiently tackle this problem, we propose \ourmethod, a meta-learning framework that uses a sparse probing dataset to train an auxiliary \textit{rater network}, which assigns utility scores to simulation samples based on their contribution to real-world validation performance, automatically guiding the active generation of a targeted synthetic dataset.
In summary, our core contributions are:
\begin{itemize}[leftmargin=*, itemsep=0pt, topsep=0pt, parsep=0pt, partopsep=0pt]
    \item We formulate \textit{Active Generative Simulation}, shifting simulation data generation from fixed distribution to controllable and targeted synthesis.
    \item We introduce \ourmethod, a meta-learning framework that trains an auxiliary rater network via meta-gradients to discover high-utility data-generation distributions within a user-specified parameter space, without manual parameter alignment.
    \item We show that \ourmethod outperforms uniform generation and offline retrieval, achieving a $150\%$ relative improvement in real-world full-task success over uniform generation while matching the performance of manual expert alignment.
    \item We demonstrate scalability and generalization of \ourmethod: proxy-curated datasets transfer to fine-tuning $\pi_{0.5}$, and \ourmethod can exploit partially aligned data when exact task alignment is unavailable.
\end{itemize}

%% file: sections/2_related_works.tex
\section{Related Works}
\textbf{Generative simulation.}
Recent progress in generative modeling~\cite{nvidia2024edify3d, chen2023fantasia3d}, motion synthesis~\cite{mandlekar2023mimicgen, cheng2024nodtamp, xue2025demogen}, and procedural environment generation~\cite{Deitke2022ProcTHORLa, liang2024eurekaverse} has enabled automated simulation pipelines that scale synthetic robot data~\cite{wang2023robogen, katara2024gen2sim, huagensim2, torne2024reconciling, dai2024automated, tian2025interndataa1pioneeringhighfidelitysynthetic, li2025momagen, gong2025anytask, saxena2024cdm}. These systems can generate assets, tasks, scenes, and demonstrations with limited human intervention. However, most synthesize feasible data passively without optimizing which specific samples best improve a downstream policy. In contrast, \ourmethod leverages a small set of target-task data to actively guide simulation data generation toward high-utility parameter regions.

\textbf{Sim-and-real alignment and policy co-training.}
To leverage simulation data for real-robot deployment, policies must overcome the sim-to-real gap~\cite{tobin2017domain, peng2018sim, chebotar2019closing, 9562091, ren2023adaptsim, memmel2024asid, ramos2019bayessim, wei2025simandrealcotraining, torne2024reconciling, maddukuri2025simandreal, cheng2025generalizable}. Domain randomization perturbs visual and physical properties to encourage robustness~\cite{tobin2017domain,peng2018sim,chebotar2019closing}, while system identification tunes simulator parameters to match real-world dynamics or appearance~\cite{ramos2019bayessim,9562091,memmel2024asid}. These approaches assume there exist high-fidelity alignments, but tighter matching does not necessarily translate to better policy learning when residual gaps remain~\cite{ren2023adaptsim}. Recent sim-and-real co-training methods instead mix simulated and real demonstrations, showing strong empirical gains~\cite{wei2025simandrealcotraining, torne2024reconciling, maddukuri2025simandreal, cheng2025generalizable, lei2026mechanisticanalysis}. \ourmethod builds on this co-training paradigm but treats the generative simulator as a controllable distribution, actively optimizing data synthesis rather than manually tuning the simulator to match the real world.

\textbf{Data attribution and retrieval in robotics.}
Recent work shows that heterogeneous robot datasets can cause negative transfer on specific tasks~\cite{agia2025cupid, chen2025curating, dass2025datamil, memmel2024strap, lin2024flowretrieval, hejna2025robot, xie2025data}. Retrieval and attribution methods address this by selecting useful samples from existing datasets using embedding similarity~\cite{memmel2024strap, du2023behavior, lin2024flowretrieval, xie2025data}, rollout outcomes~\cite{agia2025cupid}, or validation loss~\cite{dass2025datamil}. However, these methods are fundamentally limited by the coverage of the static prior dataset and cannot synthesize new target-relevant data. \ourmethod is complementary: it uses utility estimation to generate new simulation data tailored to the target task.

%% file: sections/3_preliminary.tex
\section{The Active Generative Simulation Problem}
We formalize \emph{Active Generative Simulation} as the problem of optimizing a simulator's data generation distribution to maximize downstream policy performance on a real-world target task. Rather than treating the simulator as a fixed data source, we formulate it as a controllable generator parameterized by spatial, visual, and task-level variables, such as object placements, camera poses, and textures. By treating data generation itself as an optimization problem, our goal is to actively allocate the simulation budget toward the parameter regions most valuable for sim-and-real co-training.

\subsection{Generative Simulation and Co-Training}

We denote the generative simulator as $G: \Phi \to \Xi$, where $\Phi$ is the space of data generation parameters and $\Xi$ is the space of expert trajectories. Given $\phi \in \Phi$, the simulator generates a demonstration $\xi \sim G(\phi)$. The parameter vector $\phi$ controls variables such as task identity, object placements, camera configuration, and visual appearance. 
 In our real-world experiments, $\Phi$ includes task mode, table texture (3D), camera yaw (1D), and object placements (4D). We assume that $\Phi$, its ranges, and the underlying generation pipeline are specified by the user; our goal is to optimize the data-generation distribution \emph{within} this space rather than to design the space itself.

Given a target parameter distribution $\mathcal{S}_\phi$, the simulator generates $\mathcal{D}_S = \{\xi^{(i)} \sim G(\phi^{(i)}) \mid \phi^{(i)} \sim \mathcal{S}_\phi\}_{i=1}^{N_S}$, which augments a small real dataset $\mathcal{D}_R = \{\xi_R^{(i)}\}_{i=1}^{N_R}$ with $N_S \gg N_R$. A policy $\pi_\theta$ is then trained with the standard sim-and-real co-training loss:
$
    \mathcal{L}_{\text{co-train}}(\theta; \mathcal{D}_{R}, \mathcal{D}_{S})
    = \alpha \cdot \mathcal{L}(\theta; \mathcal{D}_{S})
    + (1 - \alpha) \cdot \mathcal{L}(\theta; \mathcal{D}_{R})
$,
where $\mathcal{L}(\theta; \mathcal{D}) = \mathbb{E}_{\xi \sim \mathcal{D}} [ \sum_{(o, a) \in \xi} \ell(\pi_\theta(o), a)]$ is the behavior cloning loss and $\alpha \in [0, 1]$ is a mixing ratio~\cite{maddukuri2025simandreal}. The central challenge is therefore: \emph{How should we choose $\mathcal{S}_\phi$ so that the generated simulation data most improves the policy performance on the target task?}

\subsection{Problem Formulation: Bilevel Optimization}

Our goal is to choose a target parameter distribution $\mathcal{S}_\phi^*$ such that a policy co-trained on real data and simulated data sampled from $\mathcal{S}_\phi^*$ maximizes real-world task performance. Let $J_R(\pi)$ denote the real-world performance metric (e.g., success rate). We formulate this objective as:
\begin{equation} \label{eq: active generative simulation problem}
\begin{aligned}
    \mathcal{S}_\phi^* = \arg\max_{\mathcal{S}_\phi \subset \Phi} \, J_R(\pi_{\theta^*}), \qquad
    \text{s.t.} \quad \theta^* = \arg\min_\theta \, \mathcal{L}_{\text{co-train}}(\theta; \mathcal{D}_R, \{G(\phi) \mid \phi \sim \mathcal{S}_\phi\}).
\end{aligned}
\end{equation}

The upper-level objective optimizes the simulation data distribution, while the lower-level objective trains the policy with standard sim-and-real co-training. Directly solving~\eqref{eq: active generative simulation problem} is intractable because $\Phi$ is high-dimensional, real-world evaluation is costly, and the optimization is nested. In Section~\ref{sec: methods}, we propose a tractable framework to approximate this objective efficiently.

%% file: sections/4_method.tex
\begin{figure*}
    \centering
    \includegraphics[width=\linewidth]{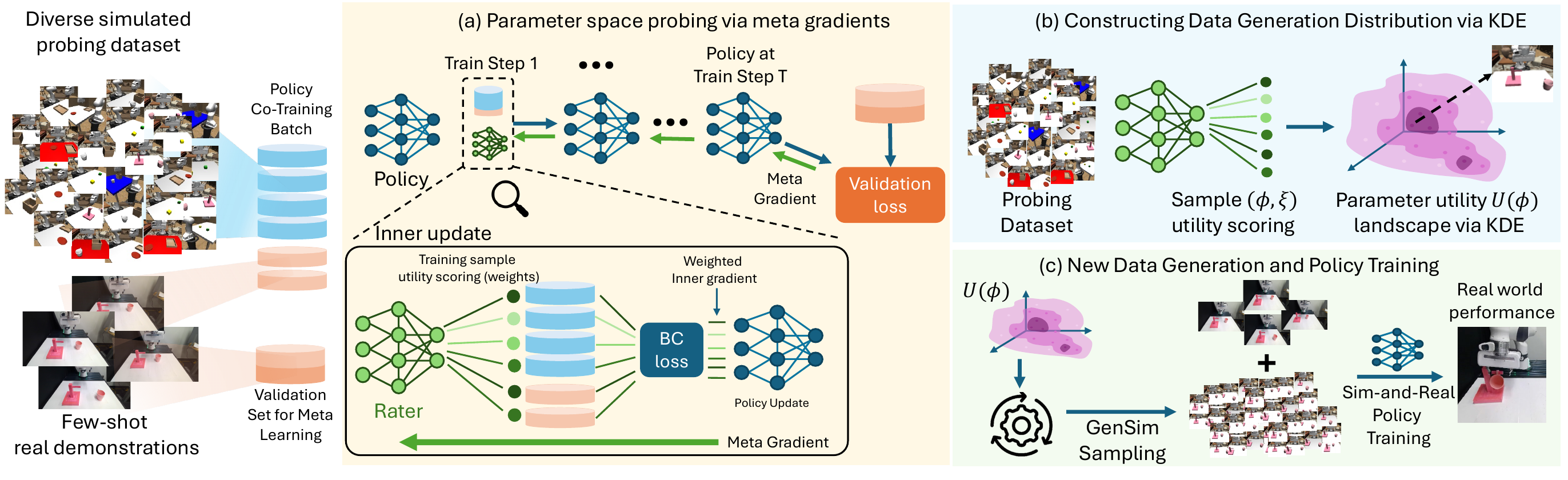}
    \caption{\textbf{\ourmethod.} We propose a meta-learning framework for Active Generative Simulation with three stages:
(a) \textbf{Rater network learning:} We meta-learn a rater network that weights simulation samples according to their contribution to policy performance on a held-out validation set of real demonstrations.
(b) \textbf{Parameter utility landscape:} We map these scores back to their corresponding simulation parameters and use kernel density estimation (KDE) to construct a probability landscape that identifies high-utility regions.
(c) \textbf{Targeted synthesis and co-training:} We sample from the constructed distribution to generate a targeted simulation corpus, then co-train the final policy on this synthetic data and few-shot real demonstrations to maximize real-world success.
}
    \label{fig:method}
\end{figure*}

\section{Active Generative Simulation by Meta Learning}\label{sec: methods}

To approximate the bilevel problem in Eq.~\eqref{eq: active generative simulation problem}, we learn an explicit approximation to the optimal target parameter distribution $\mathcal{S}_\phi^*$. Our framework has two stages: (1) learn an auxiliary \textit{rater network} that scores simulation samples by their contribution to real-task performance~\cite{engstrom2025optimizing,calian2025datarater}, and (2) convert these scores into a parameter distribution $\widehat{\mathcal{S}}_\phi$ for targeted data generation.
We assume access to a \textit{probing dataset} $\mathcal{D}_{\text{probe}} = \{(\phi_i, \xi_i)\}_{i=1}^{N_P}$, where $\phi_i$ are uniformly sampled from $\Phi$ and $\xi_i \sim G(\phi_i)$. It provides broad coverage of $\Phi$ and links each parameter setting to its generated trajectory, enabling us to estimate which regions of $\Phi$ are most useful for constructing $\mathcal{D}_S$.

\subsection{Task-Aware Data Rater Meta-Learning}
\label{sec:method-rater}

Directly optimizing over the combinatorial space $\Phi$ is intractable, so we learn a rater network $R_\eta(o, a)$ that assigns a scalar \textit{utility score} to any observation-action pair.

\textbf{Weighted Co-training (Inner Loop).} At training step $t$, we sample a simulation mini-batch $\mathcal{B}_S$ from $\mathcal{D}_{\text{probe}}$ and a real mini-batch $\mathcal{B}_R$ from $\mathcal{D}_R$ according to the co-training ratio $\alpha$. The rater computes normalized weights
$w_{t,i} = \frac{\exp{R_\eta(o_i, a_i)}}{\sum_{j \in \mathcal{B}_S \cup \mathcal{B}_R} \exp{R_\eta(o_j, a_j)}} \label{eq:softmax-weights}$
and updates the policy $\pi_\theta$ via the weighted gradient
$g_t
= \nabla_\theta \left[\sum_{i \in \mathcal{B}_S} w_{t,i} \mathcal{L}(\theta;o_i, a_i)+\sum_{j \in \mathcal{B}_R} w_{t,j} \mathcal{L}(\theta; o_j, a_j) \right],
\label{eq:inner-update}$
where $\mathcal{L}(\theta;o, a) = \ell(\pi_\theta(o), a)$ is the behavior cloning loss, and policy parameters are updated by a gradient descent algorithm as $\theta_{t+1} = \Theta\big(\theta_t,\, g_t\big)$. We apply weights to both simulation and real samples.

\textbf{Meta-Optimization (Outer Loop).} To connect training with real-world performance $J_R$, we split the real dataset into training and validation subsets and use the validation loss on $\mathcal{D}_{R,\text{val}}$ as a differentiable surrogate. The meta-objective is to find $\eta$ that minimizes the validation loss after unrolling $T$ steps of inner-loop updates:
$
\eta^\star = \arg\min_{\eta} \mathcal{L}(\theta_T(\eta); \mathcal{D}_{R,\text{val}}).
\label{eq:outer-objective}
$
We update $\eta$ by backpropagating through the $T$ inner-loop updates and applying an outer-loop gradient descent step on this validation loss.
Thus, $R_\eta$ acts as a proxy for $J_R$, assigning high scores to simulation samples whose gradients reduce prediction error on real-world expert trajectories. The inner-loop policy is used only to provide meta-gradients for learning $R_\eta$; after constructing $\widehat{\mathcal{S}}_\phi$, we train the final policy on the generated co-training dataset. The full meta-learning procedure is provided in Algorithm~\ref{alg:metagensim_rater} in Appendix~\ref{app:metagensim details}.

\subsection{Automatic Parameter Selection and Demo Generation}
\label{sec:method-kde}
The rater $R_\eta$ operates on observation-action pairs, but Eq.~\eqref{eq: active generative simulation problem} requires a distribution $\mathcal{S}_\phi$ over simulator parameters. We bridge this gap by mapping rater scores back to the probing parameters and then constructing $\widehat{\mathcal{S}}_\phi$.

\textbf{Parameter utility.}
We define a utility function $\mathcal{U}: \Phi \to \mathbb{R}$ that maps each generation parameter to its estimated contribution to real-world performance. Since the rater $R_\eta$ assigns scores to individual observation-action pairs, we estimate the utility of a parameter $\phi_i$ by aggregating the rater scores along the corresponding synthesized trajectory $\xi_i = \{(o_{i,t}, a_{i,t})\}_{t=1}^{T_i}$:
$
    \mathcal{U}(\phi_i)
    =
    \frac{1}{T_i}
    \sum_{t=1}^{T_i}
    R_\eta(o_{i,t}, a_{i,t}).
$
Optionally, this empirical mean can be replaced with a CVaR-style aggregation to provide additional robustness against high variance in the quality of synthesized $(o,a)$ pairs (see Appendix~\ref{app:metagensim details}). Since $\mathcal{D}_{\text{probe}}$ uniformly covers $\Phi$, these aggregated scores define a utility landscape used to approximate $\mathcal{S}_\phi^*$.

\textbf{Task-conditioned density estimation.} We represent $\widehat{\mathcal{S}}_\phi$ with a task-conditioned density. We decompose each $\phi = (m, x)$ into a discrete task mode $m \in \mathcal{M}$ and continuous parameters $x \in \mathbb{R}^p$ such as object poses, camera poses, and texture RGBs. We sample from $p_{\widehat{\mathcal{S}}}(m, x)$ by factorizing it as $p_{\widehat{\mathcal{S}}}(m, x) = p(m)p(x \mid m)$ and proceeding in two steps:

\begin{enumerate}[leftmargin=*, itemsep=0pt, topsep=0pt, parsep=0pt, partopsep=0pt]
    \item Discrete task selection: Let $\mathcal{I}_m$ be the indices in $\mathcal{D}_{\text{probe}}$ belonging to task $m$, and let $\bar{\mathcal{U}}_m = \frac{1}{|\mathcal{I}_m|} \sum_{i \in \mathcal{I}_m} \mathcal{U}(\phi_i)$ be the average utility of that task. We then sample $m$ with probability $p(m) \propto \exp(\bar{\mathcal{U}}_m / \tau_m)$, where $\tau_m$ is a task-level temperature coefficient.
    
    \item Continuous parameter density estimation: Conditioned on $m$, we fit a weighted KDE over $x$:
    $p(x \mid m) = \sum_{i \in \mathcal{I}_m} w_i \mathcal{K}_H(x - x_i)$,
    where $\mathcal{K}_H$ is a Gaussian kernel with bandwidth $H$ set using Scott's rule~\cite{scott2015multivariate}. Since $\mathcal{D}_{\text{probe}}$ is sparse, raw utilities can be noisy. We compute $w_i$ by first applying kernel-weighted local averaging to smooth utilities among neighboring parameters within task $m$, then applying a temperature-scaled softmax. We provide the exact weighting equations in Appendix~\ref{app:metagensim details}.
\end{enumerate}

\noindent
Sampling $\phi \sim p_{\widehat{\mathcal{S}}}$ and generating $\xi \sim G(\phi)$ yields the targeted simulation dataset $\mathcal{D}_S$ used for final co-training. Invalid samples are rejected using simulator constraints. This concentrates the generation budget in regions expected to improve real-world performance, turning the choice of $\mathcal{S}_\phi$ into a density-estimation problem.

%% file: sections/5_experiments.tex
\section{Experiments}
\label{sec:experiments}

\newcommand{\res}[2]{#1\textcolor{gray}{\scriptsize $\pm$#2}}
\newcommand{\rescount}[1]{#1\textcolor{gray}{\scriptsize/20}}
\newcommand{\resTotal}[1]{#1\textcolor{gray}{\scriptsize/80}}

In this section, we design experiments across simulation and real-world manipulation tasks to evaluate \ourmethod. Specifically, our evaluations answer four questions: \textbf{RQ1 (Efficacy):} Does \ourmethod improve sim-and-real co-training over real-only training, uniform generation, retrieval, and manual alignment? \textbf{RQ2 (Mechanism):} Does \ourmethod learn target-aware utility scores and data-generation distributions? \textbf{RQ3 (Cross-architecture transfer):} Do distributions curated with a lightweight proxy policy transfer to large VLA policies? \textbf{RQ4 (Partial alignment):} When exact sim-and-real alignment is impossible, can \ourmethod still exploit partially aligned simulation data to improve target performance?

\begin{figure*}[t]
        \centering
\includegraphics[width=1\linewidth,trim={0 0.6cm 0 0},clip]{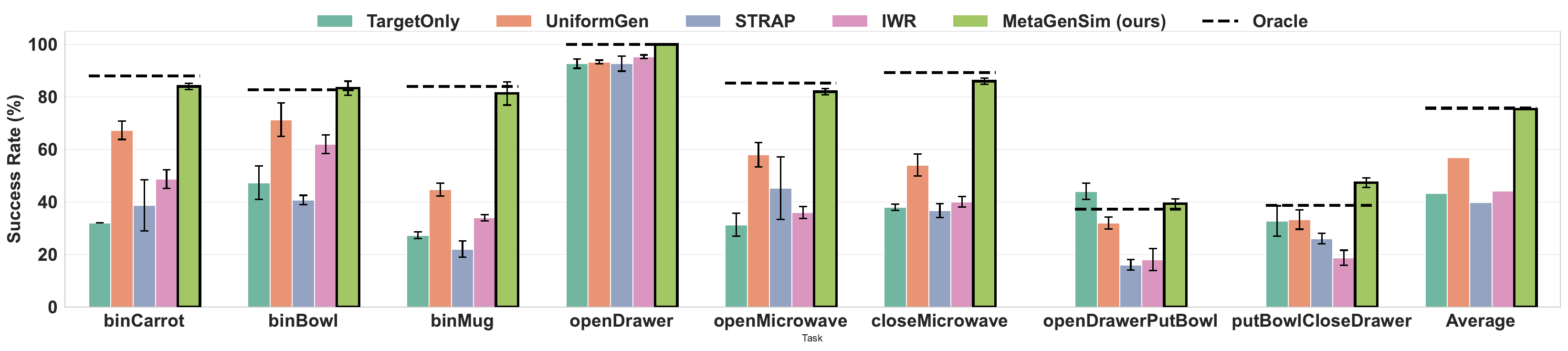}
        \caption{Sim-and-Sim Co-training performance using BC-RNN. Success bars are mean over 3 seeds, showing Standard Error of the Mean (SEM). Diffusion Policy exhibits a similar trend; due to space constraints, we include those results in Appendix~\ref{app:additional_main_results}.}
        \label{fig:bcrnn_sim_performance}
    \end{figure*}

\input{tables/sim2real_results_dp}

\textbf{Sim-and-real experiments.} We evaluate \ourmethod on four challenging real-world target tasks: \textit{stackCube}, \textit{mugBin}, \textit{bowlTray}, and \textit{mugHang}. We collect only 10 real demonstrations for each task. In simulation, the source task set contains the same four tasks, and the data generation parameters include a discrete task mode and 8 continuous dimensions: camera yaw (1D), table RGB (3D), and object placements (4D). We instantiate the generative simulator with customized MimicLabs~\cite{saxena2025mimiclabs} environments and use MimicGen~\cite{mandlekar2023mimicgen} to generate simulation demonstrations. We use BC-RNN as the inner-loop model during meta-learning, then train final policies on the datasets constructed by each method. We co-train Diffusion Policy (DP)~\cite{chi2023diffusionpolicy, chi2024diffusionpolicy} with ratio $\alpha=0.9$ following~\citet{maddukuri2025simandreal}. We evaluate the policies on a real Franka Panda robot using 20 rollouts and report Pick and Pick+Place success rates. Additional setup details are provided in Appendix~\ref{app:sim-and-real setup}--\ref{app:train and eval details}.

\textbf{Sim-and-sim experiments.} We conduct controlled experiments in simulation on a subset of MimicLabs dataset~\cite{saxena2025mimiclabs}. We use 8 target tasks, ranging from pick-and-place to long-horizon manipulation, and the data generation parameters include tasks, labs, camera pose, object placement parameters. The target and source domains use different backgrounds and table textures to simulate a cross-domain gap. We use BC-RNN as the inner-loop model during meta-learning, then co-train final BC-RNN and DP policies on the datasets constructed by each method. We report peak success rates from 50 rollout evaluations, averaged across three seeds. See Appendix~\ref{app:sim-and-sim setup}--\ref{app:train and eval details} for full details.

\textbf{Baselines.} We compare \ourmethod with the following data generation and curation strategies: (1) \uniformgen: Generate simulation data by uniformly sampling parameters from $\Phi$. (2) \oracle: Manually align data generation parameters and generate simulation data. (3) Retrieval: Adapt \STRAP~\cite{memmel2024strap} and \IWR~\cite{xie2025data} to select a subset of simulation data from \uniformgen based on similarity to the target task. Further details on the retrieval baselines are provided in Appendix~\ref{app:baselines}. We also compare with \texttt{RealOnly}, where the policy is only trained on real demonstrations.

\subsection{Efficacy of Targeted Generation (RQ1)}
\label{sec:results}

\input{figures/score_distribution/rater-score-distribution}

We compare \ourmethod against \texttt{RealOnly} and \uniformgen, offline retrieval methods (\STRAP and \IWR), na\"ive scaling of uniformly generated data, and manual alignment (\oracle).

\textbf{Significant gains over \uniformgen and \texttt{RealOnly}.} \autoref{fig:bcrnn_sim_performance} presents the BC-RNN sim-and-sim results over 8 target tasks; DP follows a similar trend and is shown in Appendix~\ref{app:additional_main_results}. Compared to \uniformgen, \ourmethod improves average policy success by 18.6 and 13.3 percentage points for BC-RNN and DP, respectively. \autoref{tab:sim2real_results_transposed_full} presents the real-world results. Policies trained only on real data struggle on most tasks, and \uniformgen can even underperform \texttt{RealOnly}, indicating that unaligned simulation data can cause negative transfer. In contrast, \ourmethod more than doubles full-task success over \uniformgen.
Appendix Table~\ref{tab:multitask_results} further shows that \ourmethod can target \textit{stackCube} and \textit{mugBin} jointly with a single rater, improving over real-only training on both tasks.

\textbf{Data quality matters more than na\"ively increasing the number of demonstrations.} To test whether the gains come simply from adding more simulation data, we scale \uniformgen by $5\times$ on \textit{stackCube} and \textit{mugBin}. As shown in Appendix Table~\ref{tab:sim2real_scaling_unif_gen}, increasing uniformly generated demonstrations from 1600 to 8000 either stalls or degrades performance. In contrast, \ourmethod achieves substantially better performance using only 800 actively generated demonstrations, showing that targeted data generation is more important than data quantity alone.

\textbf{\ourmethod outperforms retrieval from a prior simulation dataset.} Retrieval is an effective approach for curating useful data from a fixed prior dataset, and both \STRAP and \IWR improve over \uniformgen on several real-world tasks. However, retrieval is limited by the coverage of the original \uniformgen dataset. 
In contrast, \ourmethod first probes the parameter space and then generates new demonstrations near high-utility parameters. As shown in \autoref{tab:sim2real_results_transposed_full}, \ourmethod achieves substantially higher real-world co-training performance than both retrieval baselines.

\textbf{\ourmethod matches manual alignment.} Across both sim-and-sim and sim-and-real experiments, \ourmethod achieves performance on par with \oracle. For BC-RNN and DP in simulation, \ourmethod performs similarly to \oracle on average. In real-world evaluations, \ourmethod matches the total Pick+Place success of expert-aligned simulation data. These results show that \ourmethod can recover the benefits of manual alignment without tuning data generation parameters by hand.

\subsection{Mechanism Analysis: Achieving Automatic Alignment (RQ2)}
\label{sec:mechanism-analysis}

\input{figures/sim2real_tsne/figure}

We analyze how \ourmethod assigns utility scores over $\mathcal D_\text{probe}$ and converts these scores into target-aligned data generation distributions.

\textbf{\ourmethod can score high-quality data for sim-and-real co-training.} To understand what the rater network learns over $\mathcal D_\text{probe}$, \autoref{fig:rater-score-distribution} visualizes the rater score distribution for the real-world \textit{stackCube} task. The rater assigns higher scores to samples from the simulated \textit{stackCube} task and suppresses scores for misaligned tasks. The scored real dataset is shown for reference, and we observe that aligned simulation data can contain high-utility samples comparable to real data in terms of rater scores.

\textbf{\ourmethod utility scores induce target-aware generation distributions.} \autoref{fig:sim2real_tsne} shows that high-scoring simulation parameters align with the target task in the high-dimensional parameter space, and that sampling from the induced distribution yields real-task-aware simulation data. Appendix~\ref{app:additional_main_results} further visualizes the generated data distributions in Figure~\ref{fig:active-gen-data-nonstandard}, showing that task, camera yaw, table color, and object placement distributions adapt appropriately when the target setting changes.

Beyond the main experiments, Appendix~\ref{app:additional_analysis_results} shows that \ourmethod can also optimize higher-dimensional camera poses (Figure~\ref{fig:high-dim-cam}) and physics parameters for dynamic tasks (Figure~\ref{fig:physics_params}). We also provide sensitivity analyses for data budgets, co-training ratio, and KDE hyperparameters.

\subsection{Proxy-curated data transfers to VLA fine-tuning (RQ3)}
\label{sec:vla-transfer}

\input{tables/pi05_results}

Directly integrating large VLA models into the meta-learning inner loop is computationally prohibitive, so we use lightweight BC-RNN in the meta-learning loop to curate simulation datasets and then fine-tune the pretrained $\pi_{0.5}$ VLA~\cite{intelligence2025pi05visionlanguageactionmodelopenworld} on the resulting co-training data. Table~\ref{tab:pi05} reports full-task success across the four real-world tasks. \ourmethod consistently outperforms both \uniformgen and \texttt{RealOnly}, while remaining close to \oracle, indicating that the learned simulation distributions are useful beyond the proxy policy class used during meta-learning. 

\subsection{Handling Irreducible Sim-and-Real Misalignment (RQ4)}
\label{sec:skill-composition}

The previous sections show that \ourmethod can align simulation parameters to real-world tasks despite visual rendering gaps; here, we study a stronger form of misalignment where the exact target task cannot be generated directly in the simulator. We test whether \ourmethod can synthesize data mixtures that support two forms of generalization: atomic skill stitching for long-horizon behaviors and object-receptacle composition for novel real tasks.

\textbf{Atomic skill stitching for long-horizon behaviors.} We first examine long-horizon tasks whose exact unified behaviors are absent from the source task set. For \textit{openDrawerPutBowl} and \textit{putBowlCloseDrawer}, full task-level alignment is impossible; useful co-training data must instead come from related atomic tasks. As shown in \autoref{fig:bcrnn_sim_performance}, \ourmethod remains effective by using meta-gradient feedback to prioritize necessary source-task components. On \textit{putBowlCloseDrawer}, for example, \ourmethod even outperforms \oracle by generating a non-obvious mixture of \textit{binBowl}, \textit{putBowlInDrawer}, and \textit{closeDrawer}. This demonstrates that \ourmethod can curate targeted data that helps policies stitch reusable simulation skills over time under irreducible task misalignment.

\begin{wrapfigure}{r}{0.48\linewidth}
  \centering
  \includegraphics[width=\linewidth]{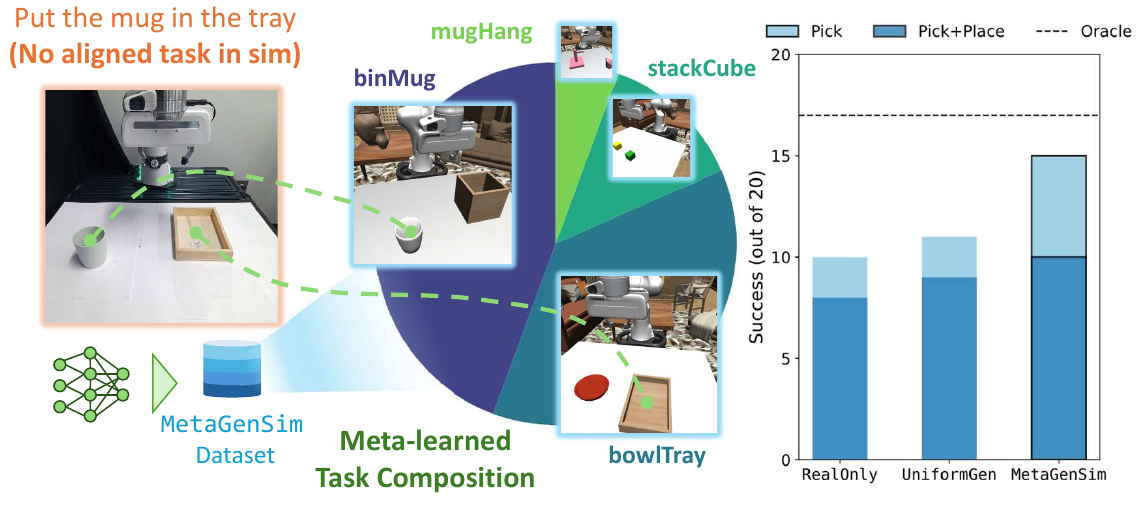}
  \caption{Generalization to compositionally novel target tasks. (Left) The real-world task \textit{mugTray} lacks a direct simulation counterpart. \ourmethod constructs a co-training dataset by identifying and prioritizing aligned sub-skills, specifically \textit{mugBin} for grasping the mug and \textit{bowlTray} for placing an object on the tray. (Right) Real-robot success counts.}
  \label{fig:mugTray}
  \vspace{-0.5cm}
\end{wrapfigure}

\textbf{Recombining Skills for Compositional Generalization.} We further evaluate whether the generated data supports generalization to the novel real-world task \textit{mugTray}, which has no exact source-task counterpart. The target task setup is unseen, but it shares the mug-grasping affordance of \textit{mugBin} and the tray-receiving affordance of \textit{bowlTray}. As shown in \autoref{fig:mugTray}, \ourmethod automatically identifies and prioritizes these related skills, generating most of its co-training data from \textit{mugBin} and \textit{bowlTray}, and achieves higher performance than \uniformgen. These results show that \ourmethod can synthesize reusable simulation-skill mixtures that enable policies to compose learned object and receptacle affordances for novel real-world target setups.

%% file: tables/sim2real_results_dp.tex
\begin{table*}[t]
    \centering
    \scriptsize %
    \setlength{\tabcolsep}{4pt}
    \caption{Sim-and-Real Co-training results for Diffusion Policy on 4 real-world tasks. Success counts are for Pick and Pick+Place stages. The best performing method (excluding \oracle) is highlighted in bold.}
    \begin{tabular}{c|cc|cc|cc|cc|cc}
        \toprule
        \textbf{Method} & \multicolumn{2}{c|}{{stackCube}} & \multicolumn{2}{c|}{{mugBin}} & \multicolumn{2}{c|}{{bowlTray}} & \multicolumn{2}{c|}{{mugHang}} & \multicolumn{2}{c}{\textbf{Total}} \\
        & Pick & Pick+Place & Pick & Pick+Place & Pick & Pick+Place & Pick & Pick+Place & Pick & Pick+Place \\
        \midrule
        \texttt{RealOnly} & \rescount{9} & \rescount{4} & \rescount{4} & \rescount{4} & \rescount{16} & \rescount{13} & \rescount{6} & \rescount{5} & \resTotal{35} & \resTotal{26} \\
        \uniformgen       & \rescount{7} & \rescount{2} & \rescount{4} & \rescount{3} & \rescount{17} & \rescount{15} & \rescount{5} & \rescount{3} & \resTotal{33} & \resTotal{23} \\
        \STRAP            & \rescount{6} & \rescount{5} & \rescount{6} & \rescount{5} & \rescount{13} & \rescount{12} & \rescount{11} & \rescount{5} & \resTotal{36} & \resTotal{27} \\
        \IWR              & \rescount{10} & \rescount{4} & \rescount{5} & \rescount{4} & \rescount{\textbf{19}} & \rescount{17} & \rescount{14} & \rescount{8} & \resTotal{48} & \resTotal{33} \\
        \activegen (ours)        & \rescount{\textbf{17}} & \rescount{\textbf{14}} & \rescount{\textbf{14}} & \rescount{\textbf{14}} & \rescount{\textbf{19}} & \rescount{\textbf{19}} & \rescount{\textbf{17}} & \rescount{\textbf{11}} & \resTotal{\textbf{67}} & \resTotal{\textbf{58}} \\
        \midrule
        \oracle           & \rescount{15} & \rescount{11} & \rescount{16} & \rescount{16} & \rescount{19} & \rescount{19} & \rescount{16} & \rescount{11} & \resTotal{66} & \resTotal{57} \\
        \bottomrule
    \end{tabular}
    \label{tab:sim2real_results_transposed_full}
\end{table*}

%% file: figures/score_distribution/rater-score-distribution.tex
\begin{wrapfigure}{r}{0.48\linewidth}
  \centering
  \vspace{-1.1cm}
  \includegraphics[width=\linewidth]{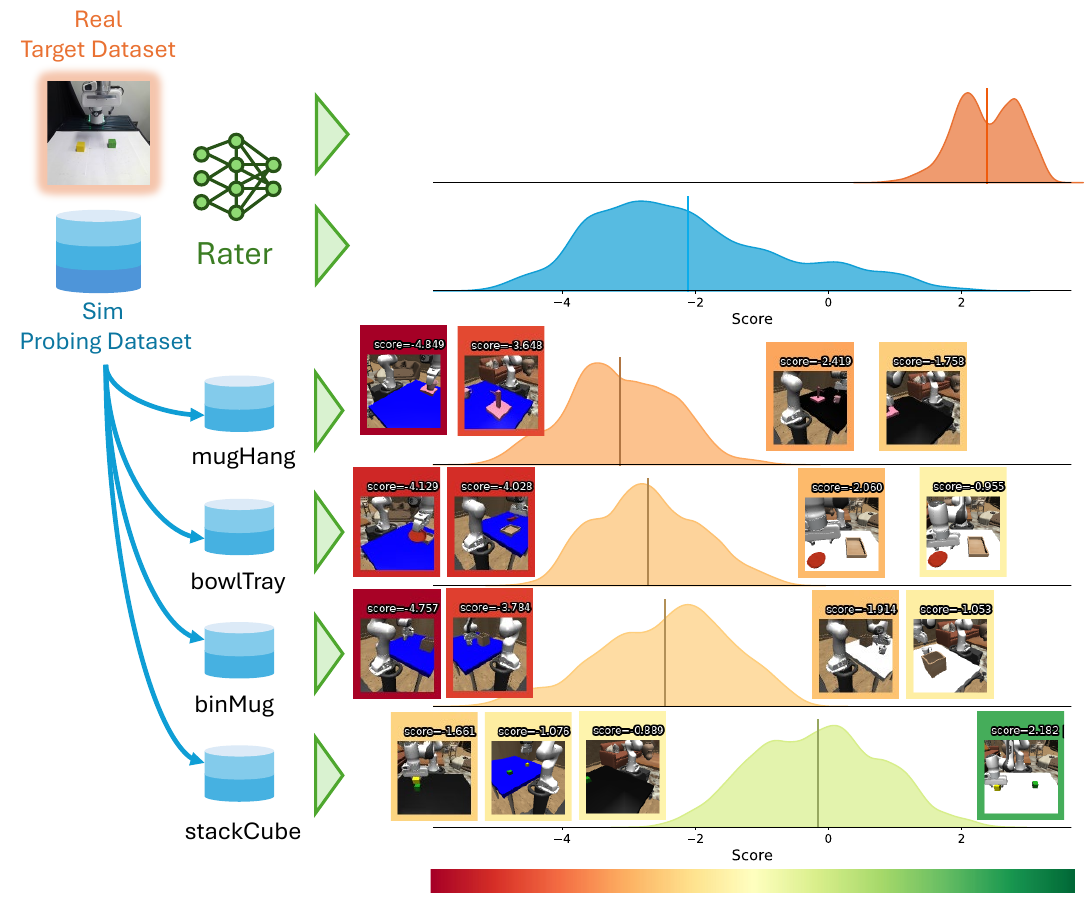}
  \caption{Rater score distribution of the probing and the real dataset ({\color{ForestGreen}green} for high, {\color{BrickRed}red} for low). The probing dataset is decomposed into four tasks with representative scored samples. The rater assigns high utility scores to the relevant \textit{stackCube} task while suppressing scores for misaligned tasks. The vertical lines represent mean scores.}
  \label{fig:rater-score-distribution}
  \vspace{-0.8cm}
\end{wrapfigure}

%% file: figures/sim2real_tsne/figure.tex
\begin{wrapfigure}{r}{0.48\linewidth}
    \vspace{-0.3cm}
    \centering
    \includegraphics[width=\linewidth]{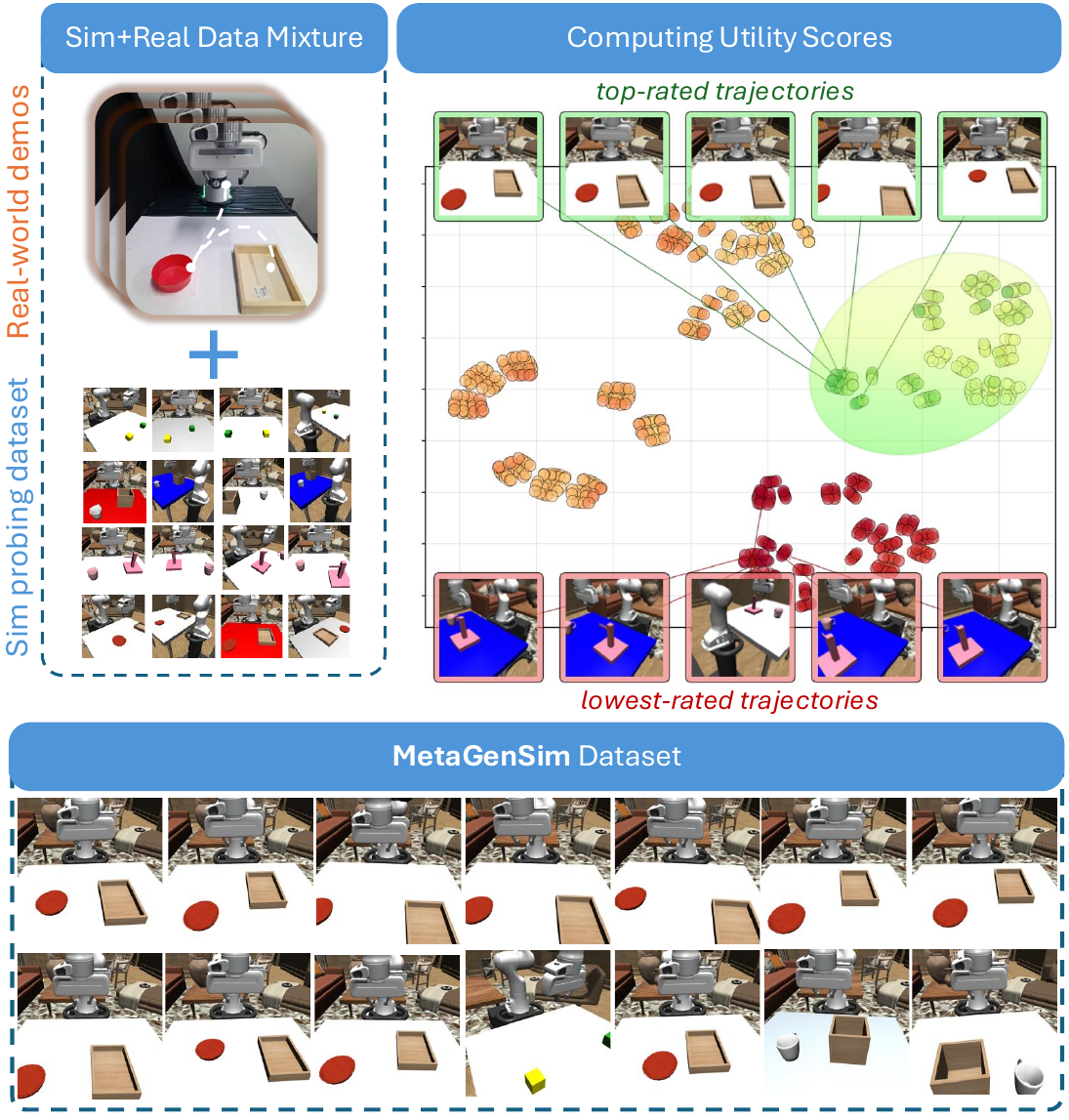}
    \caption{\textbf{\ourmethod for real-world tasks.}
(Top-left) We meta-learn the rater network using 10 real demonstrations and an 800-demo simulated probing dataset spanning 4 tasks.
(Top-right) The trained rater assigns utility scores to probing data. We visualize their 12-D task configurations (table color, camera yaw, object spatial arrangements, and one-hot vector indicating tasks) using t-SNE, colored by utility scores ({\color{ForestGreen}green}: high; {\color{BrickRed}red}: low). High-scoring trajectories match the target task in visual and spatial arrangements, while low-scoring ones are misaligned.
(Bottom) We sample from the induced parameter distribution to generate a large-scale, target-aware simulation dataset.}
    \label{fig:sim2real_tsne}
    \vspace{-0.9cm}
\end{wrapfigure}

%% file: tables/pi05_results.tex
\begin{wraptable}{r}{{0.49\linewidth}}
    \centering
    \scriptsize
    \setlength{\tabcolsep}{2pt}
    \vspace{-6mm}
    \caption{Full-task success counts for fine-tuning the pretrained $\boldsymbol{\pi_{0.5}}$ VLA.}
    \begin{tabular}{c|cccc|c}
    \toprule
    \textbf{Method} & \textit{stackCube} & \textit{mugBin} & \textit{bowlTray} & \textit{mugHang} & \textbf{Total} \\
    \midrule
    \texttt{RealOnly} & \rescount{1} & \rescount{6} & \rescount{12} & \rescount{0} & \resTotal{19} \\
    \uniformgen & \rescount{2} & \rescount{12} & \rescount{13} & \rescount{3} & \resTotal{30} \\
    \ourmethod & \rescount{\textbf{11}} & \rescount{\textbf{18}} & \rescount{\textbf{18}} & \rescount{\textbf{5}} & \resTotal{\textbf{52}} \\
    \midrule
    \oracle & \rescount{13} & \rescount{19} & \rescount{17} & \rescount{7} & \resTotal{56} \\
    \bottomrule
    \end{tabular}
    \label{tab:pi05}
    \vspace{-4mm}
\end{wraptable}

%% file: sections/6_appendix.tex
\section{Overview}

The Appendix is organized as follows:
\begin{itemize}
    \item \textbf{Sim-and-Real Experiments Setup} (Appendix~\ref{app:sim-and-real setup}): This section details the real-world target tasks and the data generation parameters.
    \item \textbf{Sim-and-Sim Experiments Setup} (Appendix~\ref{app:sim-and-sim setup}): This section describes the simulation experiments, including the construction of domain gaps using MimicLabs and the selection of target tasks.
    \item \textbf{Details on \ourmethod} (Appendix~\ref{app:metagensim details}): This section specifies the architecture of the Rater Network, the full meta-learning algorithm, and the generation hyperparameters used in our experiments.
    \item \textbf{Policy Training and Evaluation} (Appendix~\ref{app:train and eval details}): This section lists the hyperparameters, training and evaluation procedures for the policies.
    \item \textbf{Retrieval Baselines} (Appendix~\ref{app:baselines}): This section details the implementation of the retrieval baselines (\STRAP and \IWR).
    \item \textbf{Additional Main Experiment Results} (Appendix~\ref{app:additional_main_results}): This section provides additional real-world visualizations, data-scaling results, and sim-and-sim policy results.
    \item \textbf{Additional Analysis and Ablations} (Appendix~\ref{app:additional_analysis_results}): This section includes studies on physics parameters, multi-task targets, higher-dimensional camera poses, sensitivity to data budgets and co-training ratio, and KDE hyperparameters.
\end{itemize}

\section{Sim-and-Real Experiments Setup}
\label{app:sim-and-real setup}

\begin{figure}[t]
    \centering
    \includegraphics[width=\linewidth]{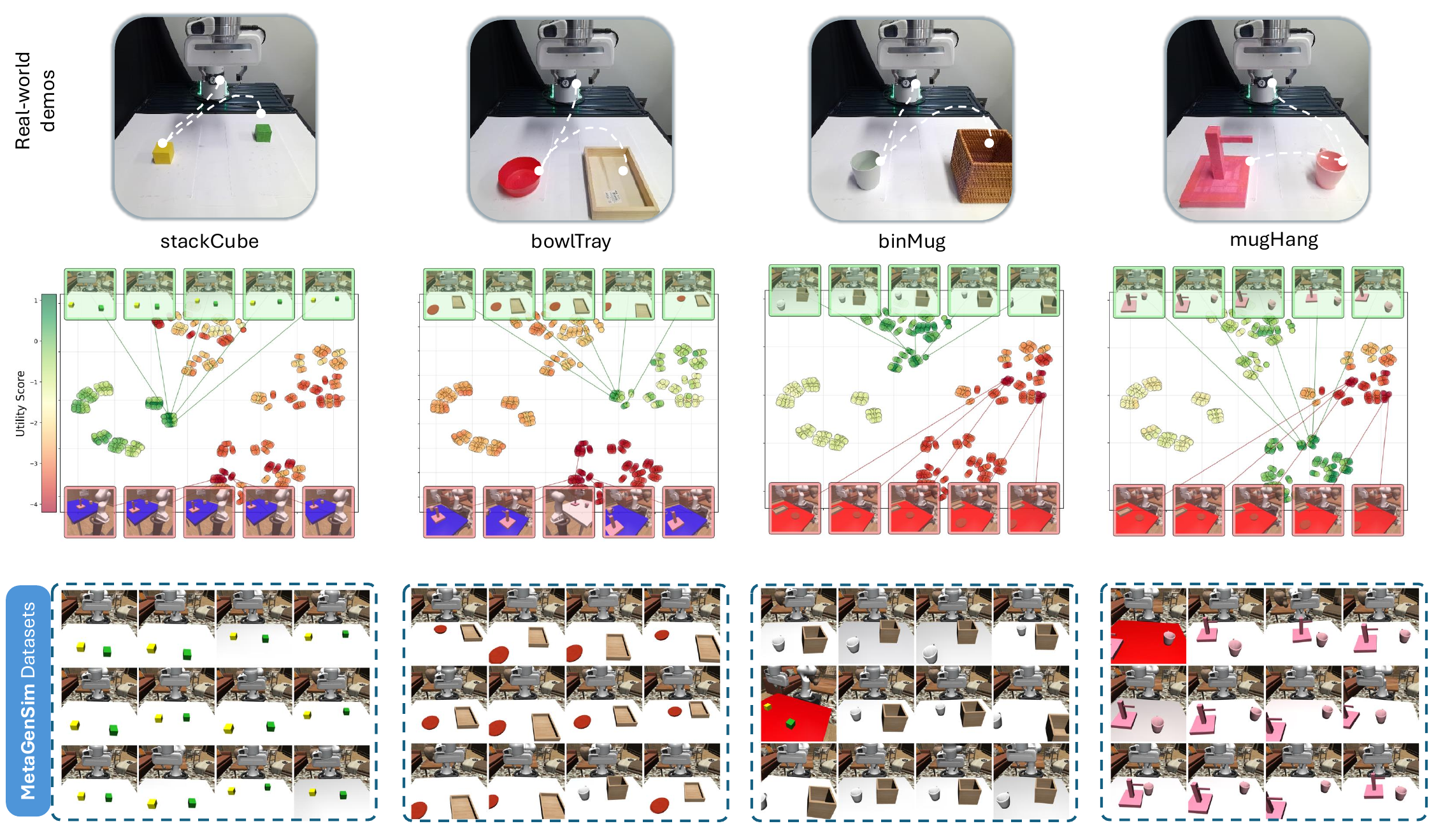}
    \caption{\textbf{\ourmethod for Real-World Tasks and Parameter Space Probing.} \textbf{(Top)} The four real-world target tasks considered in our main experiments. \textbf{(Middle)} We use the trained rater network to compute utility scores for each trajectory in the probing dataset. We show t-SNE plots where the 12-D task configurations for each trajectory (3-D table color, 1-D camera yaw, 2-D spatial arrangements for two objects, and a 4-D one-hot vector representing task) are fit to 2-D, and colored as per their utility score ({\color{ForestGreen}green} for high, {\color{BrickRed}red} for low). The top-scored trajectories can be observed to match the visual and spatial arrangements of the real task, while the lowest-rated trajectories differ significantly in either aspect. \textbf{(Bottom)} Samples from the final \ourmethod dataset, generated by sampling from the induced data generation parameter distribution.}
    \label{fig:tsne-all-tasks}
\end{figure}

\textbf{Target Tasks}. To validate the effectiveness of our method under realistic scenarios, we evaluate \ourmethod and baselines on four challenging real-world target tasks: \textit{stackCube}, \textit{mugBin}, \textit{bowlTray}, and \textit{mugHang}. A visualization of 4 target tasks can be found in Figure~\ref{fig:tsne-all-tasks}. These tasks require high-precision manipulation and object alignment. To enable fine-grained performance analysis, we decouple the evaluation into two stages: \textit{Pick} and \textit{Place}. We report the success counts for Pick and Pick+Place in Table~\ref{tab:sim2real_results_transposed_full}. Detailed task descriptions and their respective success criteria are listed below:
\begin{itemize}
    \item \textit{stackCube}: The robot must pick up a yellow cube and stack it on top of a green cube.
    \begin{itemize}
        \item \textbf{Pick Success}: The robot successfully grasps and lifts the yellow cube.
        \item \textbf{Place Success}: The yellow cube remains stably stacked on the green cube after the robot releases its gripper.
    \end{itemize}

    \item \textit{mugBin}: The robot must pick up a mug and place it inside a bin.
    \begin{itemize}
        \item \textbf{Pick Success}: The mug is successfully grasped and lifted.
        \item \textbf{Place Success}: The mug remains inside the bin after the robot releases the gripper.
    \end{itemize}

    \item \textit{bowlTray}: The robot must pick up a bowl and place it onto a tray.
    \begin{itemize}
        \item \textbf{Pick Success}: The bowl is successfully grasped and lifted.
        \item \textbf{Place Success}: The bowl is placed flatly inside the tray after the gripper is released.
    \end{itemize}

    \item \textit{mugHang}: The robot must pick up a mug and hang it on a mug tree branch by its handle.
    \begin{itemize}
        \item \textbf{Pick Success}: The mug is successfully grasped and lifted.
        \item \textbf{Place Success}: The mug remains hanging on the mug tree branch after the robot releases the gripper.
    \end{itemize}
\end{itemize}

\textbf{Target Dataset}: We collected 10 demonstrations for each target task via teleoperation using a Meta Quest 3 headset and controller. These real-world demonstrations form the target dataset, denoted as $\mathcal{D}_R$.

\begin{wrapfigure}{r}{0.48\linewidth}
  \centering
  \vspace{-0.3cm}
  \includegraphics[width=1.0\linewidth]{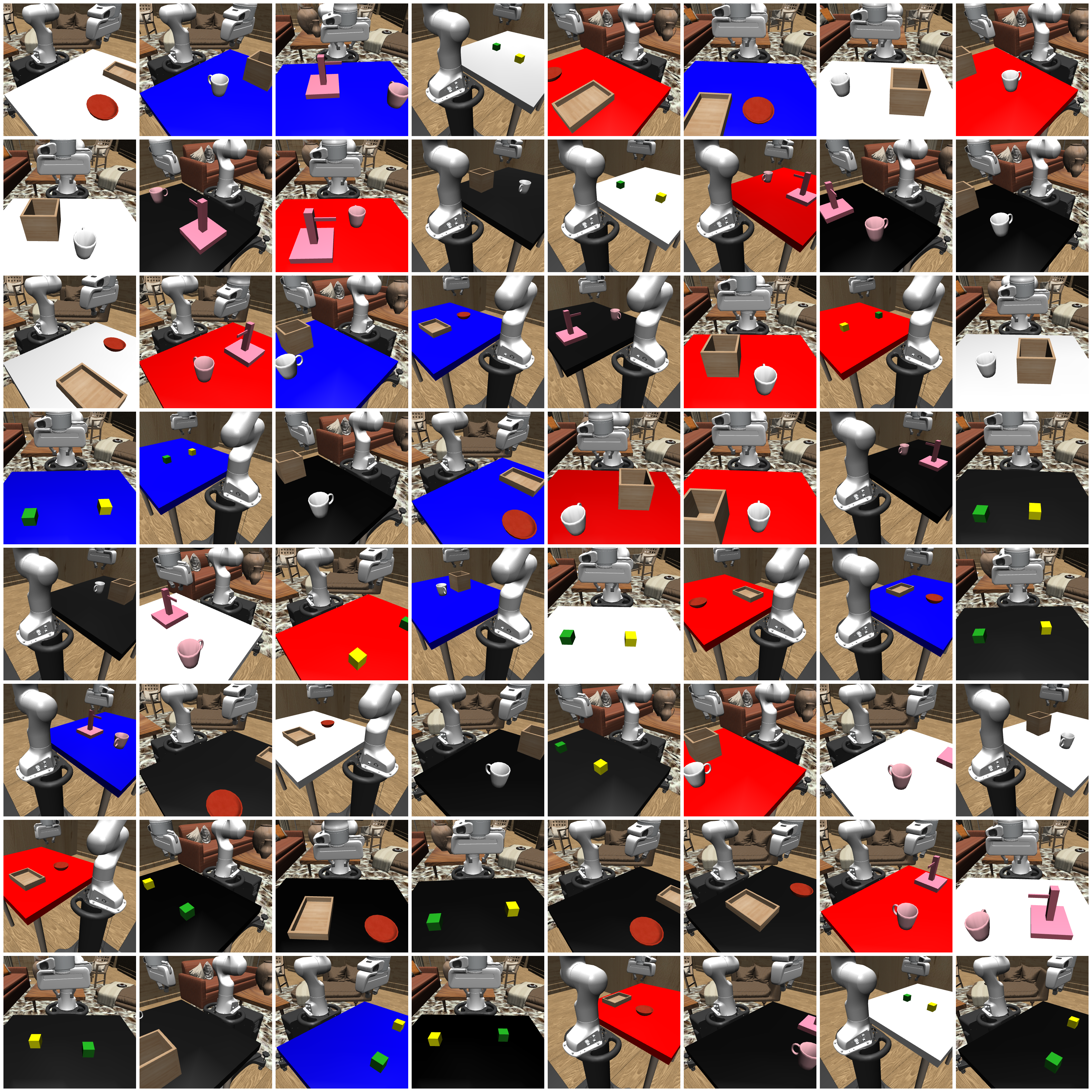}
  \caption{\textbf{Visualization of initial frames from the probing dataset for the sim-and-real co-training experiments.} The dataset provides broad coverage across all tasks and exhibits a diverse range of camera poses, table textures, and object placements, ensuring comprehensive exploration of the parameter space.}
  \label{fig:probing_dataset_examples}
  \vspace{-0.5cm}
\end{wrapfigure}

\textbf{Data Generation Parameters}: As discussed in Section~\ref{sec:experiments}, we parameterize the generative simulator using Task (Discrete), Camera Pose (Continuous, 1D), Table Texture (Continuous, 3D), and Object Placements (Continuous, 4D). We still use MimicLabs as our generative simulator, but instead of using MimicLabs datasets, we customize the environment to match the real table height, camera intrinsics, and camera pointing position. Also, we choose one specific lab (Lab3) for our experiments and do not consider lab as a data generation parameter. As can be seen in Figure~\ref{fig:probing_dataset_examples}, the background of the simulation environments are very different from the real environment.

Although the parameter space $\Phi$ is continuous, the generative simulator imposes validity constraints. For instance, object and receptacle positions cannot overlap to avoid collisions. Furthermore, the underlying demonstration generator, MimicGen~\cite{mandlekar2023mimicgen}, requires distinct source demonstrations for different spatial arrangements (e.g., swapping objects from left to right requires separate source demos). To satisfy these constraints when generating $\mathcal{D}_{\text{probe}}$ and the \uniformgen dataset, we sample parameters from pre-defined disjoint regions. Specifically, object positions are drawn from two distinct configuration settings: (Object 1 on the left, Object 2 on the right) and (Object 1 on the right, Object 2 on the left), where ``left'' and ``right'' denote specific $15\text{cm} \times 20\text{cm}$ rectangular zones. Additionally, camera yaw is sampled from 5 disjoint angular bins, and table textures are sampled from regions near black, white, blue, and red. When using \ourmethod to actively generate data, we do not restrict sampling to these disjoint regions. Instead, we sample continuous values directly from the Gaussian kernels. For object positions, we apply a simple rejection sampling step to prevent objects from being too close to each other or out of the workspace.

\textbf{Baselines and Datasets}. We generate simulation demonstrations via MimicGen~\cite{mandlekar2023mimicgen}. We compare \ourmethod against the following baselines, controlling for data budgets where applicable:

\begin{itemize}
    \item \textbf{\ourmethod (Ours):} We first generate a probing dataset $\mathcal{D}_{\text{probe}}$ of 800 demos (Visualization in Figure~\ref{fig:probing_dataset_examples}), which is used alongside $\mathcal{D}_R$ to train the rater network (details in Appendix~\ref{app:metagensim details}). We then set a generation budget of 800 demos for the active generation phase. Thus, the final \ourmethod dataset consists of 800 demos sampled from the learned distribution.
    
    \item \textbf{\oracle:} This dataset contains 800 demos generated with \emph{expert-aligned} parameters: the correct task, camera extrinsics from the real-world calibration, manually tuned object reset ranges, and manually selected table colors.
    
    \item \textbf{\uniformgen:} To ensure a fair comparison of total simulation samples, \uniformgen has access to the 800 probing demos plus an additional 800 uniformly generated demos, totaling 1600 demos.
    
    \item \textbf{\STRAP and \IWR:} These retrieval baselines select a subset of data from the 1600-demo \uniformgen pool. Further implementation details for retrieval-based methods are provided in Appendix~\ref{app:baselines}.
\end{itemize}
We use the same source demos in MimicGen to generate all above simulation datasets.

\textbf{Generation Hyperparameters}. For all real-world tasks, we use the same default hyperparameters when constructing the generation distribution: task temperature $0.5$, trajectory-score temperature $0.05$, and KDE multiplicative coefficients $0.6$ for camera yaw, $0.2$ for table RGB, and $0.8$ for object positions. The sensitivity of performance to these hyperparameters is analyzed in Appendix~\ref{app:additional_analysis_results}.

After generating the simulation datasets, we co-train Diffusion Policy~\cite{chi2023diffusionpolicy, chi2024diffusionpolicy} on a mixture of the generated simulation data and the real-world data $\mathcal{D}_R$. We utilize a co-training ratio of $\alpha=0.9$, which is determined by empirical findings in~\cite{maddukuri2025simandreal}. For details on training and evaluation procedures, please refer to Appendix~\ref{app:train and eval details}.

For the real-world \textit{stackCube} setting with up-and-down object placements (shown in Figure~\ref{fig:active-gen-data-nonstandard}), we treat the task as fixed rather than a data generation parameter. Consequently, for this specific experiment, the probing dataset consists exclusively of samples from the \textit{stackCube} task.

\section{Sim-and-Sim Experiments Setup}
\label{app:sim-and-sim setup}

We conduct controlled experiments in simulation to evaluate the efficacy of our meta-learning algorithm in identifying simulation parameters that result in environments aligned with the target task. These experiments are based on a subset of the MimicLabs dataset~\cite{saxena2025mimiclabs}.

\textbf{Domain Gap Construction.} To simulate the sim-to-real gap, we manually create domain shifts using the distinct ``lab" environments provided in MimicLabs. The dataset contains 8 different labs, each featuring unique background scenes and table textures. For each target task, we designate one lab as the \textbf{target domain} (from which the few-shot target data is drawn) and the remaining 7 labs as the \textbf{source domains}. The generative simulator is constrained to generate data only within the source domains. Consequently, a domain gap is strictly enforced, as the visual appearance (background and texture) of the generated data always differs from the target task.

\begin{figure}[t]
  \centering
  \includegraphics[width=0.7\linewidth]{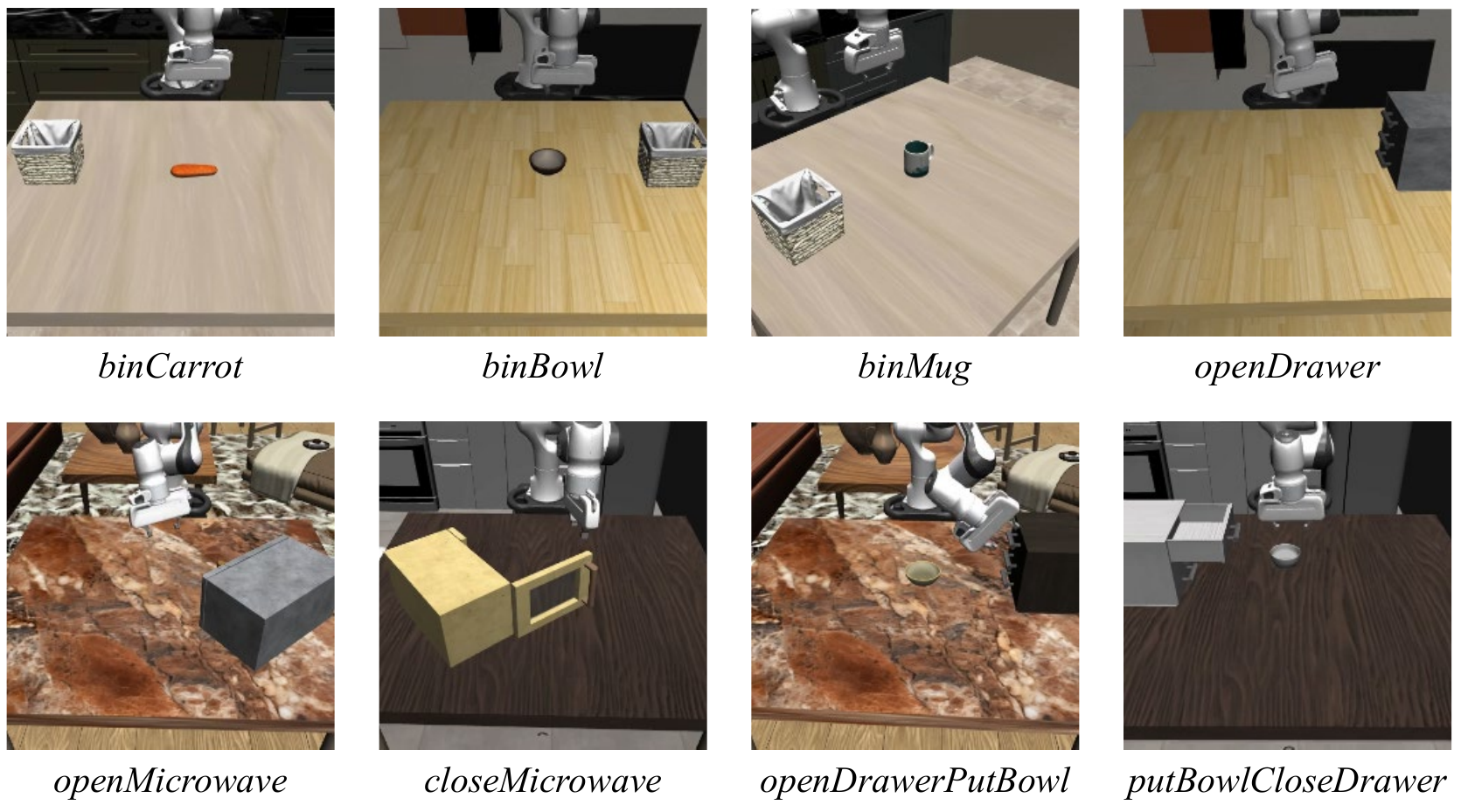}
  \caption{Target tasks for sim-and-sim co-training experiments.}
  \label{fig:sim-and-sim_setup}
\end{figure}

\textbf{Target Tasks.} As shown in Figure~\ref{fig:sim-and-sim_setup}, we select 8 target tasks for our evaluation, ranging from short-horizon pick-and-place to long-horizon multi-stage manipulation:
\begin{itemize}
    \item \textit{binCarrot}: The robot picks up a carrot and places it into a bin.
    \item \textit{binBowl}: The robot picks up a bowl and places it into a bin.
    \item \textit{binMug}: The robot picks up a mug and places it into a bin.
    \item \textit{openDrawer}: The robot pulls a drawer open.
    \item \textit{openMicrowave}: The robot opens a microwave door.
    \item \textit{closeMicrowave}: The robot closes a microwave door.
    \item \textit{openDrawerPutBowl}: A long-horizon task where the robot must first open a drawer and then place a bowl inside it.
    \item \textit{putBowlCloseDrawer}: A long-horizon task where the robot must place a bowl into a drawer and then close the drawer.
\end{itemize}

\textbf{Target Dataset}: We use 10 demonstrations for each target task, with the exception of \textit{openDrawerPutBowl}, which utilizes 20 demonstrations.

\textbf{Data Generation Parameters.} For the source data generation, we consider a parameter space defined by the following axes:
\begin{itemize}
    \item {Atomic Tasks (Discrete):} We consider 8 atomic source tasks: \textit{binCarrot}, \textit{binBowl}, \textit{binMug}, \textit{openDrawer}, \textit{closeDrawer}, \textit{openMicrowave}, \textit{closeMicrowave}, and \textit{putBowlInDrawer}. Note that while the first six target tasks have direct counterparts in the source tasks, the long-horizon target tasks do not. We expect \ourmethod to automatically identify useful atomic sub-tasks (e.g., identifying \textit{openDrawer} and \textit{putBowlInDrawer} as necessary stages for the \textit{openDrawerPutBowl} target task).
    \item {Labs (Discrete):} The 7 available source labs.
    \item {Camera Pose (Continuous):} The camera pose is represented in spherical coordinates, pointing towards the table center.
    \item {Object Spatial Arrangement (Continuous):} The reset position of the manipulated objects (e.g., mug, bowl).
    \item {Receptacle Spatial Arrangement (Continuous):} The reset position of the receptacle (e.g., bin, drawer).
\end{itemize}

The MimicLabs dataset structure inherently groups continuous parameters into bins. Camera positions are grouped into 5 bins (each with varying polar and azimuthal angles. A visualization of camera pose ranges can be found in \cite[Figure 6]{saxena2025mimiclabs}), object reset positions are grouped into 3 bins (Left, Right, Center), and receptacle positions into 2 bins (Left, Right). Combined with the atomic tasks and labs, this results in approximately 800 distinct parameter bins for the source data. For \ourmethod, $\mathcal{D}_{\text{probe}}$ contains 1600 uniformly generated demonstrations, and the data generation budget is 1600 demonstrations. 

The primary objective of the sim-and-sim experiments is to verify the Rater Network's ability to effectively score aligned data (Section~\ref{sec:method-rater}). Therefore, we apply a simplified version of \ourmethod for these specific experiments. Instead of constructing a continuous probability density function (as in Section~\ref{sec:method-kde}), we leverage the pre-existing bins described above. We rank the parameter bins according to the CVaR scores of the probing samples and generate data from the top 2\% of high-scoring bins. The $\mathcal D_\text{probe}$ and \uniformgen datasets are generated uniformly across all bins, while \STRAP and \IWR retrieve data from the \uniformgen dataset.

It is important to emphasize that this bin-ranking approach is used only for the sim-and-sim setting to validate the scoring mechanism. The Sim-and-Real experiments (detailed in Appendix~\ref{app:sim-and-real setup}) utilize the complete version of \ourmethod. In the real-world setting, we do not group parameters into bins; instead, we perform continuous density estimation (KDE) over the individual data generation parameters to construct and sample from a smooth utility landscape.

After the source datasets are generated using different methods, we co-train BC-RNN and Diffusion Policy (DP)~\cite{chi2023diffusionpolicy, chi2024diffusionpolicy} on the mixture of the generated/retrieved source dataset and the few-shot target dataset.

\section{Details on \ourmethod}
\label{app:metagensim details}

\textbf{Rater Network Architecture.} The Rater network is designed to assign a scalar score to observation-action pairs. It consists of an encoder followed by a MLP head. The encoder utilizes a ResNet18 backbone, identical to the one used in the policy architecture (see Appendix~\ref{app:train and eval details}). However, unlike the policy, the Rater Network receives both observations and actions as inputs. Specifically, the action vector is concatenated with the low-dimensional observation vector before being processed. The combined features are then passed through an MLP with two hidden layers of size (256, 256) to output the final scalar score.

\input{algorithms/alg}

\textbf{Meta-Learning Implementation.} 
The meta-learning procedure is detailed in Algorithm~\ref{alg:metagensim_rater}. Following~\cite{calian2025datarater}, we employ the Adam optimizer~\cite{kingma2017adammethodstochasticoptimization} for both the inner (policy update) and outer (rater update) loops. We utilize the differentiable Adam implementation from TorchOpt~\cite{JMLR:TorchOpt} to enable higher-order meta-gradient backpropagation. We set the inner learning rate to $1e-4$ (for the policy) and the outer learning rate to $1e-5$ (for the rater). For real-world experiments, we use $T=5$ inner gradient steps and a batch size of 64 for both the training and validation batches; this configuration fits in a single NVIDIA A40 GPU. For simulation experiments, we found that $T=2$ inner steps is sufficient to learn an effective rater.

We observed that using BC-RNN as the inner loop policy yields high-quality active generation results, even when the final downstream policy is a Diffusion Policy. We found that using Diffusion Policy directly within the meta-learning inner loop resulted in a ``flat" rater score distribution, preventing the network from effectively distinguishing aligned parameters. We hypothesize that this is due to the stochastic noise inherent in the diffusion training, though further investigation is left for future work. Consequently, we use BC-RNN as the inner policy for all experiments.

\textbf{Importance of Co-Training the Rater.} Empirically, we found it essential to input both simulation and real-world data into the rater network during the inner loop updates (as shown in Algorithm~\ref{alg:metagensim_rater}). If the rater is trained solely on the simulated probing dataset, it fails to correctly identify aligned parameters, likely due to the lack of grounding in the real-world data distribution.

\textbf{Differences from DataRater~\cite{calian2025datarater}.}
While our approach is inspired by \cite{calian2025datarater}, there are three key distinctions:
\begin{itemize}
    \item {Domain Application:} \cite{calian2025datarater} was originally proposed for curating data in the language domain (LLMs). We adapt the formulation specifically for the continuous domain of robotics and control.
    \item {Cross-Domain Scoring:} \cite{calian2025datarater} focuses on filtering out low-quality in-domain data (where validation data is from the same domain). In contrast, our algorithm solves a \textit{cross-domain} problem: we score \emph{simulation} data based on its utility for a \emph{real-world} target task.
    \item {Simplified Meta-Learning:} We do not utilize multiple inner models or periodic re-initialization of inner models. We found that a single inner policy model, no re-initialization, is sufficient for identifying high-utility regions in our setting. We do not use the bilevel gradient optimization techniques which is used to reduce RAM usage in \cite{calian2025datarater}. 
    
\end{itemize}

\textbf{CVaR score}
To ensure the policy learns from the most successful segments of a demonstration, rather than being penalized by transient recovery behaviors, we define $\mathcal{U}(\phi)$ using an Upper Conditional Value-at-Risk (CVaR): 
\begin{equation}
\mathcal{U}(\phi_i) = \mathbb{E} [s \mid s \geq q_{1-\rho}], \quad s \in {R_\eta(o,a)}_{(o,a) \in \xi_i}
\end{equation}
where $q_{1-\rho}$ is the $(1-\rho)$-quantile of scores in trajectory $\xi_i$. In practice, using mean is also good enough but CVaR makes the proposal robust to the high variance inherent in automated expert generators. Because $\mathcal{D}_{\text{probe}}$ covers the parameter space uniformly, this functional serves to directly transition the rater's evaluation of trajectories into a utility landscape over generative parameters.

\textbf{Local utility smoothing and KDE weights.}
For continuous parameter density estimation, raw utilities $\mathcal{U}(\phi_i)$ can be noisy because $\mathcal{D}_{\text{probe}}$ sparsely samples $\Phi$. We therefore smooth utilities within each discrete task mode $m$ using kernel-weighted local averaging, i.e., a Nadaraya-Watson estimator. For $i \in \mathcal{I}_m$, where $\mathcal{I}_m$ indexes probing samples with task mode $m$, we compute
\begin{equation}
\tilde{\mathcal{U}}(\phi_i)
=
\frac{
\sum_{j \in \mathcal{I}_m}
\exp\!\left(-\frac{1}{2}\lVert x_i - x_j\rVert_{\Sigma}^{2}\right)
\mathcal{U}(\phi_j)
}{
\sum_{j \in \mathcal{I}_m}
\exp\!\left(-\frac{1}{2}\lVert x_i - x_j\rVert_{\Sigma}^{2}\right)
},
\label{eq:appendix-smoothing}
\end{equation}
where $\lVert x_i - x_j\rVert_{\Sigma}^2 = (x_i - x_j)^\top \Sigma^{-1}(x_i - x_j)$ and diagonal $\Sigma$ normalizes different parameter dimensions to comparable scales. The KDE mixture weights are then obtained with a temperature-scaled softmax:
\begin{equation}
w_i
=
\frac{\exp(\tilde{\mathcal{U}}(\phi_i) / \tau_x)}
{\sum_{j \in \mathcal{I}_m}\exp(\tilde{\mathcal{U}}(\phi_j) / \tau_x)},
\label{eq:appendix-temp-scale}
\end{equation}
where $\tau_x > 0$ controls the concentration of the continuous-parameter distribution. Lower $\tau_x$ concentrates sampling near high-utility probes, while higher $\tau_x$ produces broader exploration.

\section{Policy Training and Evaluation}
\label{app:train and eval details}

\textbf{Simulation Policy Training}.
We employ BC-RNN~\cite{mandlekar2021matters} and Diffusion Policy~\cite{chi2023diffusionpolicy} as our imitation learning algorithms. Both architectures utilize a ResNet18 visual backbone. We also condition the policies on language instructions using FiLM layers in the visual backbone. All policies are trained for 500 epochs, with 500 gradient steps per epoch. Optimization is performed using the Adam optimizer~\cite{kingma2017adammethodstochasticoptimization} with a learning rate of $10^{-4}$.
The policy inputs consist of low-dimensional observations (end-effector position, quaternion, and gripper position), RGB images (captured by agentview and wrist-mounted cameras), and language embeddings. The low-dimensional observations are processed by a MLP and concatenated with image features. Original input images of resolution $84\times84$ are randomly cropped to $76\times76$ during training. The policy output consists of the delta position and rotation for the end-effector, as well as the gripper position. Hyperparameters used for policy training are reported in Table~\ref{tab:policy_hyperparams}.

\textbf{Simulation Policy Evaluation}. 
We evaluate policy checkpoints every 50 epochs, performing 50 rollouts per evaluation. For reporting, we select the peak success rate observed during training. All experiments are repeated across three random seeds, with peak success rates reported independently for each seed. In Figures~\ref{fig:bcrnn_sim_performance} and~\ref{fig:dp_sim_performance}, the bar heights represent the mean success rate across the three seeds, while the error bars indicate the Standard Error of the Mean (SEM).

\begin{table}[h]
    \centering
    \caption{Policy hyperparameters for BC-RNN and Diffusion Policy.}
    \vspace{1mm}
    \label{tab:policy_hyperparams}
    \begin{minipage}[t]{0.45\linewidth}
        \centering
        \textbf{BC-RNN}
        \vspace{1mm}

        \begin{tabular}{ll}
        \toprule
        batch size & 64 \\
        sequence length & 10 \\
        optimizer & Adam \\
        learning rate & 1e-4 \\
        image encoder & ResNet18 w/ FiLM \\
        image resolution & $(84, 84)$ \\
        \bottomrule
        \end{tabular}
    \end{minipage}
    \hfill
    \begin{minipage}[t]{0.50\linewidth}
        \centering
        \textbf{Diffusion Policy}
        \vspace{1mm}

        \begin{tabular}{ll}
        \toprule
        batch size & 64 \\
        observation horizon ($T_o$) & 2 \\
        action horizon ($T_a$) & 8 \\
        prediction horizon ($T_p$) & 16 \\
        diffusion method & DDPM \\
        optimizer & Adam \\
        learning rate & 1e-4 \\
        image encoder & ResNet18 w/ FiLM \\
        image resolution & $(84, 84)$ \\
        \bottomrule
        \end{tabular}
    \end{minipage}
\end{table}

\textbf{Real Policy Training}.
The training procedure for real-world policies mirrors the simulation setup, with a few key distinctions. We exclusively utilize Diffusion Policy for all real-world experiments. Additionally, we apply a Spatial Softmax layer~\cite{levine2016end} to the output of the visual encoders. The policy input observations remain consistent with the simulation settings, with the exception of the wrist camera image, which is omitted as it is not available on our real-world hardware. The policy output consists of the absolute position and rotation of the end-effector, along with the gripper position.

\textbf{Real Policy Evaluation}. 
To ensure a rigorous and fair comparison, we predefine 20 fixed pairs of reset positions for the object (e.g., yellow cube, mug) and the receptacle (e.g., green cube, bin). These evaluation reset configurations are distinct from the reset positions used during demonstration collection, yet lie within the same range. We evaluate every policy on these exact 20 configurations (resulting in 20 rollouts per evaluation). To enable fine-grained performance analysis, we decompose each target task into two stages (Pick and Pick+Place) and report the success counts for the two stages in~\autoref{tab:sim2real_results_transposed_full}.

\input{tables/sim2real_scaling_unif_gen}

\section{Retrieval Baselines}
\label{app:baselines}
We consider the retrieval baselines \STRAP and \IWR to investigate whether selectively curating data from a static, uniformly generated simulation dataset can enhance real-world performance when co-trained with real data.

\textbf{\STRAP Implementation.} We utilize DINOv2~\cite{oquab2023dinov2} feature embeddings to perform retrieval from the \uniformgen dataset. Specifically, we extract sub-trajectories with a minimum length of 20 steps and retrieve the top-100 matches based on feature similarity.

\textbf{\IWR implementation}. We reproduce this baseline following~\citet{xie2025data}. We first train VAE on both the simulation \uniformgen dataset and the real data, and retrieve the top $10\%$ data for co-training.

\begin{figure*}[t]
  \centering
  \begin{minipage}[t]{0.48\linewidth}
    \centering
    \includegraphics[width=\linewidth]{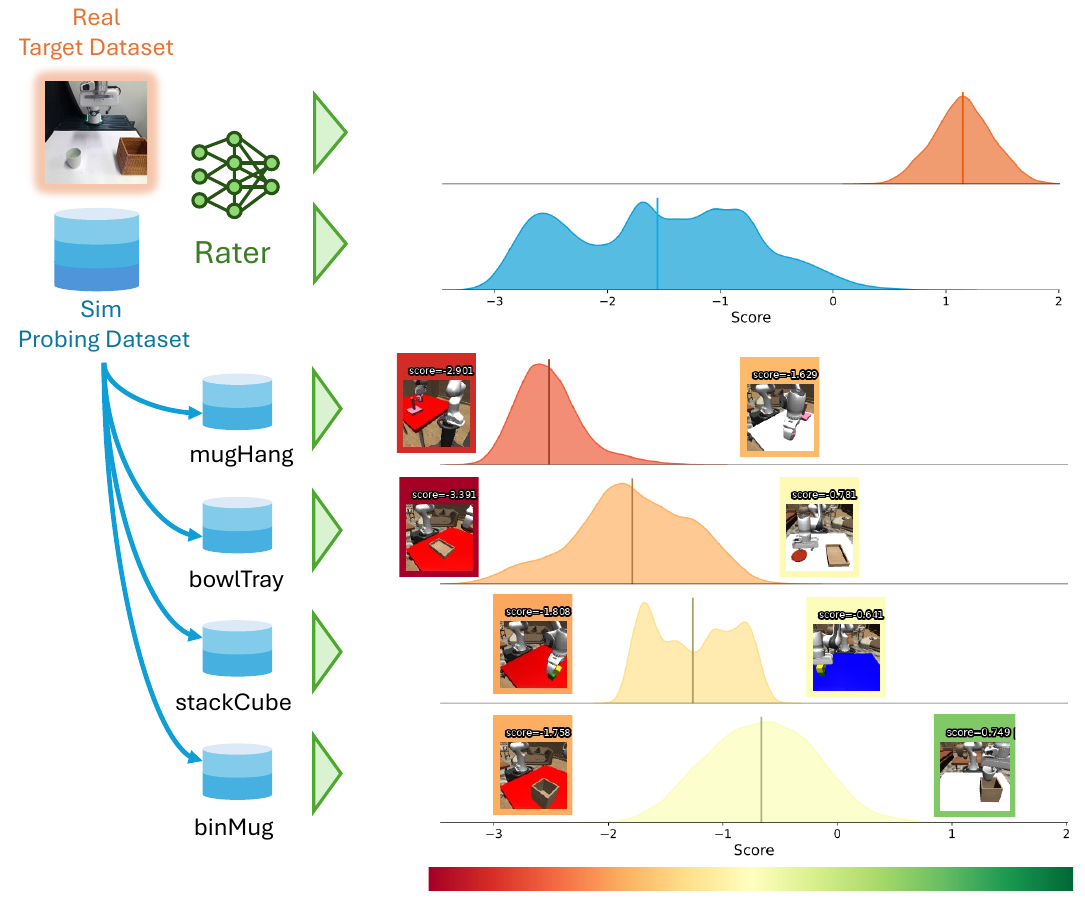}
    \vspace{-1mm}
    {\small (a) \textit{mugBin}}
  \end{minipage}
  \hfill
  \begin{minipage}[t]{0.48\linewidth}
    \centering
    \includegraphics[width=\linewidth]{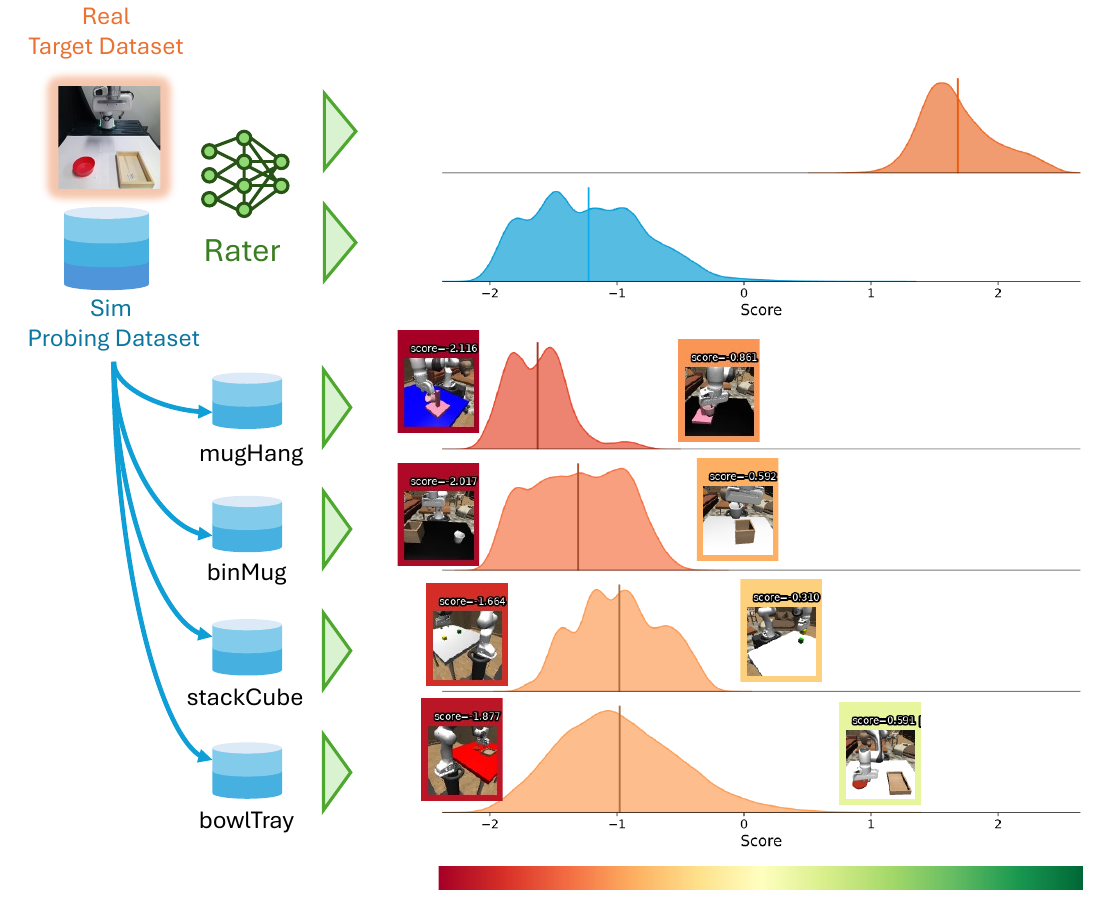}
    \vspace{-1mm}
    {\small (b) \textit{bowlTray}}
  \end{minipage}
  \vspace{2mm}

  \begin{minipage}[t]{0.48\linewidth}
    \centering
    \includegraphics[width=\linewidth]{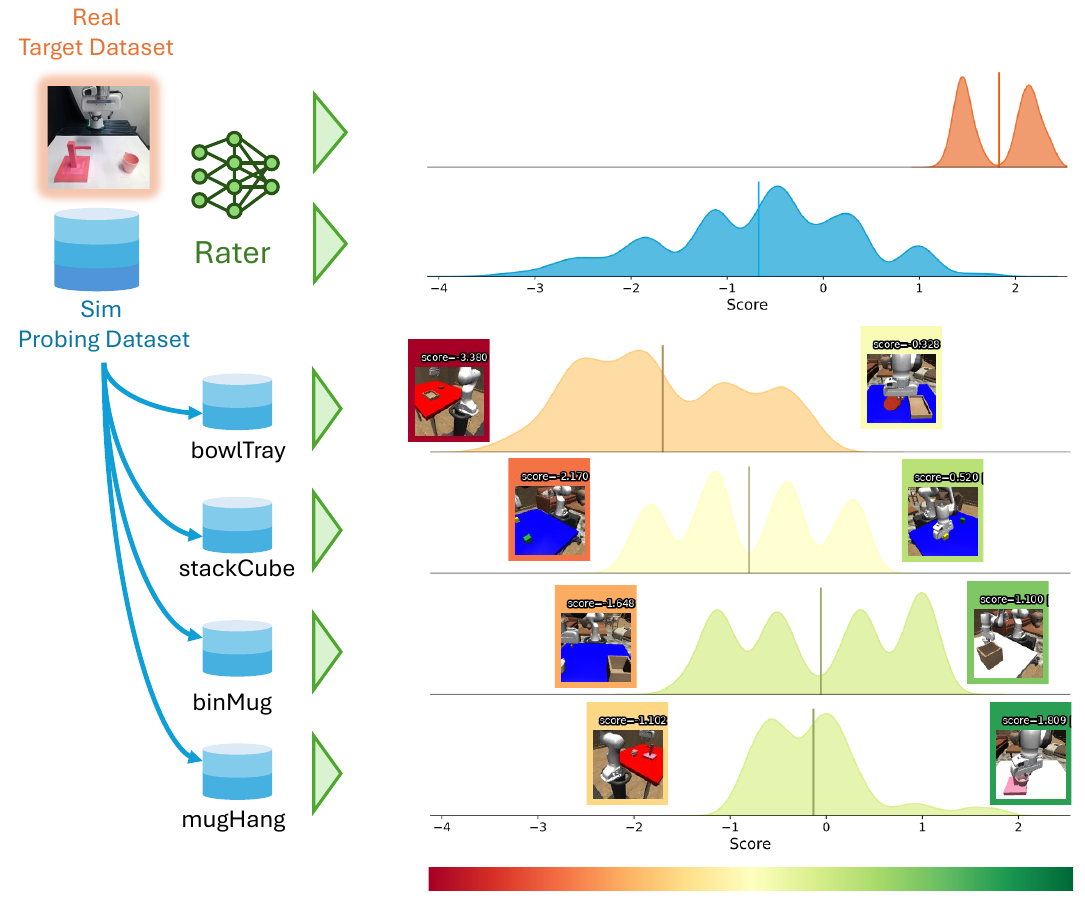}
    \vspace{-1mm}
    {\small (c) \textit{mugHang}}
  \end{minipage}
  \caption{Rater score distributions for additional real-world target tasks. The aligned simulation task generally receives higher scores and a heavier upper tail than misaligned tasks.}
  \label{fig:rater-score-distribution-additional}
\end{figure*}

\section{Additional Main Experiment Results}
\label{app:additional_main_results}

\textbf{Additional real-world rater score distributions.}
We provide visualizations of the rater score distributions for the \textit{mugBin}, \textit{bowlTray}, and \textit{mugHang} tasks in Figure~\ref{fig:rater-score-distribution-additional} (the corresponding figure for the \textit{stackCube} task is presented in Figure~\ref{fig:rater-score-distribution} of the main paper). Consistent with our main results, the rater score distribution for the aligned task subset exhibits a higher mean value and a heavier upper tail compared to misaligned tasks, confirming that the rater network consistently assigns higher utility to simulation data that aligns with the target real-world task.

\textbf{Scaling up \uniformgen.} We conduct additional \uniformgen experiments for \textit{stackCube} and \textit{mugBin}, where we co-train policies with 8000 uniform demonstrations instead of 1600. As shown in Table~\ref{tab:sim2real_scaling_unif_gen}, increasing the number of uniformly generated demonstrations either leaves performance nearly unchanged or degrades it. In contrast, \ourmethod achieves substantially better performance with only 800 actively generated demonstrations.

\begin{wrapfigure}{r}{0.48\linewidth}
  \centering
  \vspace{-0.5cm}
  \includegraphics[width=\linewidth]{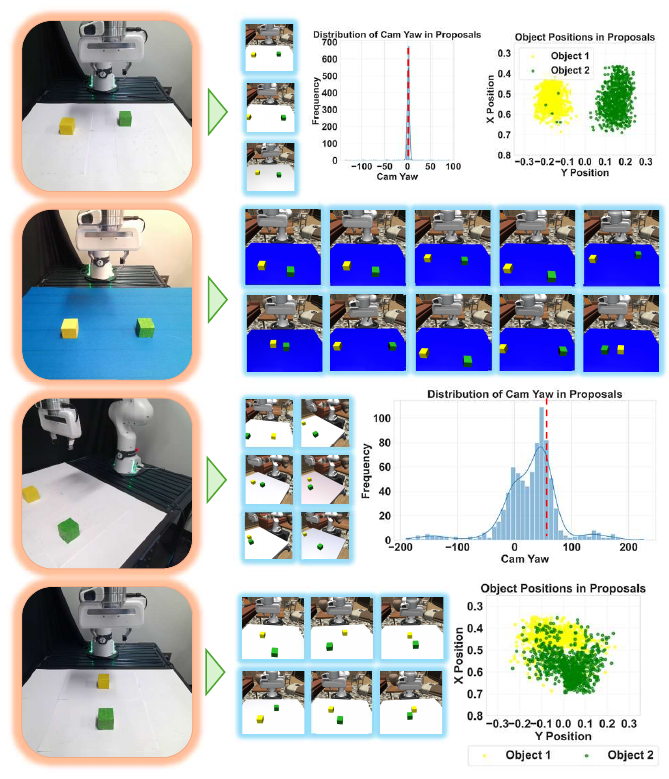}
  \caption{\ourmethod-generated data for different real-world target-task variants. Top row shows the standard setting used in our main experiments, including image samples and the distributions of camera yaw and object placements; the dashed red line denotes the ground-truth camera yaw obtained via calibration. The second row shows generated data for \textit{stackCube} with a blue table, where \ourmethod predominantly selects blue table textures. The third row considers \textit{stackCube} under a different camera pose, and the camera-yaw distribution shifts accordingly. The fourth row shows \textit{stackCube} with up-and-down object placements, where the generated placement distribution correspondingly shifts to up-and-down configurations.}
  \label{fig:active-gen-data-nonstandard}
  \vspace{-1.5cm}
\end{wrapfigure}

\textbf{Additional generated-distribution visualizations.}
In Figure~\ref{fig:tsne-all-tasks}, we visualize the scored parameters by projecting the high-dimensional parameter space into 2-D using t-SNE~\cite{maaten2008visualizing}. We observe that \ourmethod successfully identifies parameters that align with the real-world target task, assigning them higher utility scores. This scoring induces a data generation distribution that is concentrated in these aligned regions. Figure~\ref{fig:active-gen-data-nonstandard} further shows that the generated task, camera yaw, table color, and object placement distributions adapt appropriately when the target setting changes.

\begin{figure*}[t]
        \centering
\includegraphics[width=1\linewidth,trim={0 0.6cm 0 0},clip]{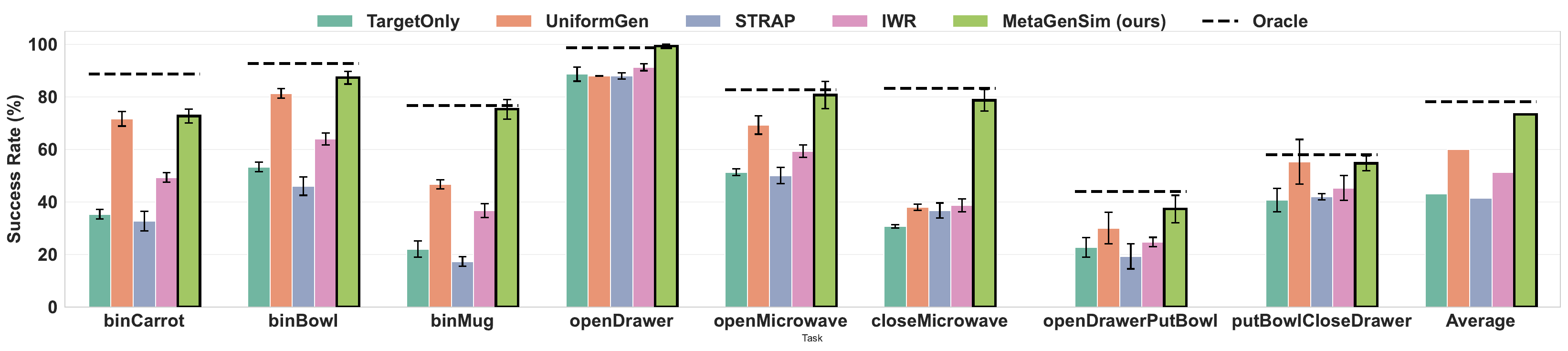}
        \caption{Sim-and-Sim Co-training performance using Diffusion Policy. Success bars are mean over 3 seeds, showing Standard Error of the Mean (SEM). The overall trend is similar to the BC-RNN results in Figure~\ref{fig:bcrnn_sim_performance}: \ourmethod consistently improves over \uniformgen and remains close to \oracle.}
        \label{fig:dp_sim_performance}
    \end{figure*}

\textbf{Diffusion Policy results for sim-and-sim experiments.}
Figure~\ref{fig:dp_sim_performance} reports the corresponding sim-and-sim results when the final policy is a Diffusion Policy. The trend is consistent with the BC-RNN results in the main paper: \ourmethod improves over \uniformgen on average and remains close to \oracle, indicating that the learned data-generation strategy is not specific to the BC-RNN policy class used in the meta-learning inner loop.

\section{Additional Analysis and Ablations}
\label{app:additional_analysis_results}

\begin{figure*}[t]
  \centering
  \includegraphics[width=0.8\linewidth]{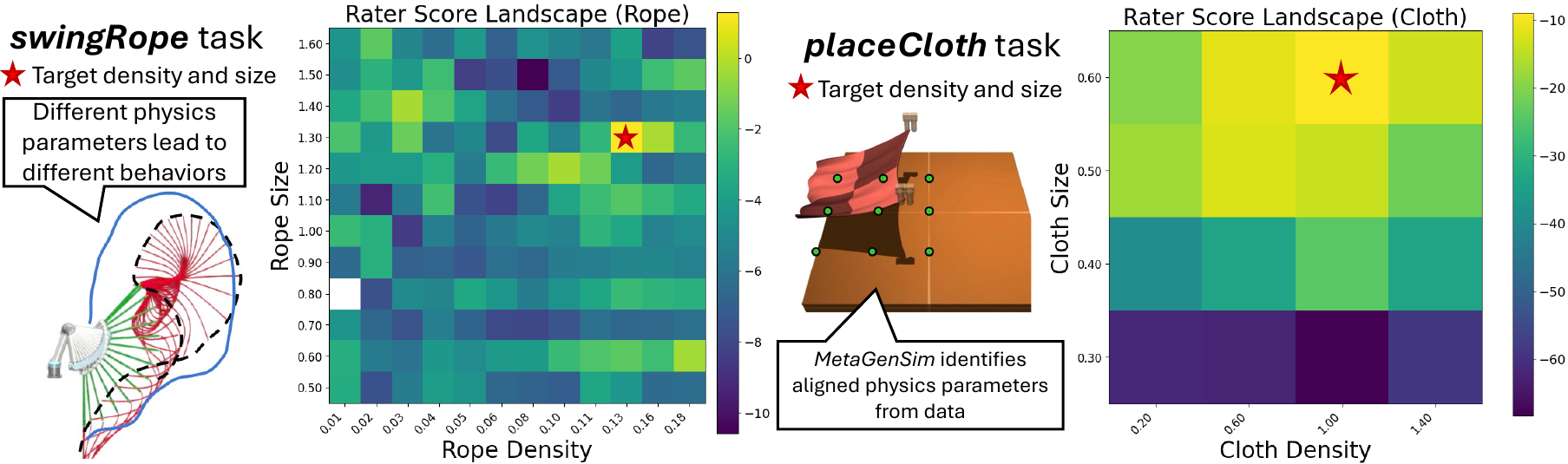}
  \caption{\ourmethod for dynamic tasks and physics parameters. Given target demonstrations for a specific rope or cloth configuration, \ourmethod identifies the aligned density and size parameters.}
  \label{fig:physics_params}
\end{figure*}

\textbf{Dynamic tasks and physics parameters.}
While our main experiments focus on quasi-static manipulation and visual/geometric parameters, the same formulation can optimize physics parameters for dynamic tasks. We evaluate \ourmethod on the dynamic \textit{swingRope} and \textit{placeCloth} tasks from IRP~\cite{Chi2022IterativeRP}. Using the official datasets, we construct candidate datasets with different rope/cloth densities and sizes, where different parameters require distinct expert actions. Given target demonstrations for a specific rope or cloth configuration, \ourmethod identifies the aligned parameter combination, as shown in Figure~\ref{fig:physics_params}. This suggests that \ourmethod can extend beyond visual and spatial variables to physics-parameter distributions.

\begin{wraptable}{r}{{0.48\linewidth}}
    \centering
    \scriptsize
    \setlength{\tabcolsep}{5pt}
    \caption{Multi-task real-world co-training results. \ourmethod is trained to target \textit{mugBin} and \textit{stackCube} simultaneously and substantially improves over target-only training on both tasks.}
    \begin{tabular}{lcc}
        \toprule
        \textbf{Task} & \texttt{TargetOnly} & \ourmethod \\
        \midrule
        \textit{mugBin}-Pick & \rescount{5} & \rescount{14} \\
        \textit{mugBin}-Pick+Place & \rescount{5} & \rescount{14} \\
        \textit{stackCube}-Pick & \rescount{7} & \rescount{17} \\
        \textit{stackCube}-Pick+Place & \rescount{3} & \rescount{17} \\
        \bottomrule
    \end{tabular}
    \label{tab:multitask_results}
\end{wraptable}

\textbf{Multi-task setting.}
To evaluate whether \ourmethod can handle multiple target tasks simultaneously, we train a single rater using target demonstrations from both \textit{stackCube} and \textit{mugBin}. The learned generation distribution produces a mixture containing $65\%$ \textit{stackCube}, $27\%$ \textit{mugBin}, and $8\%$ other tasks. The resulting co-trained policy improves over target-only training on both tasks, as shown in Table~\ref{tab:multitask_results}, achieving $17/20$ full-task successes on \textit{stackCube} and $14/20$ on \textit{mugBin}.

\begin{figure*}[t]
  \centering
  \includegraphics[width=0.8\linewidth]{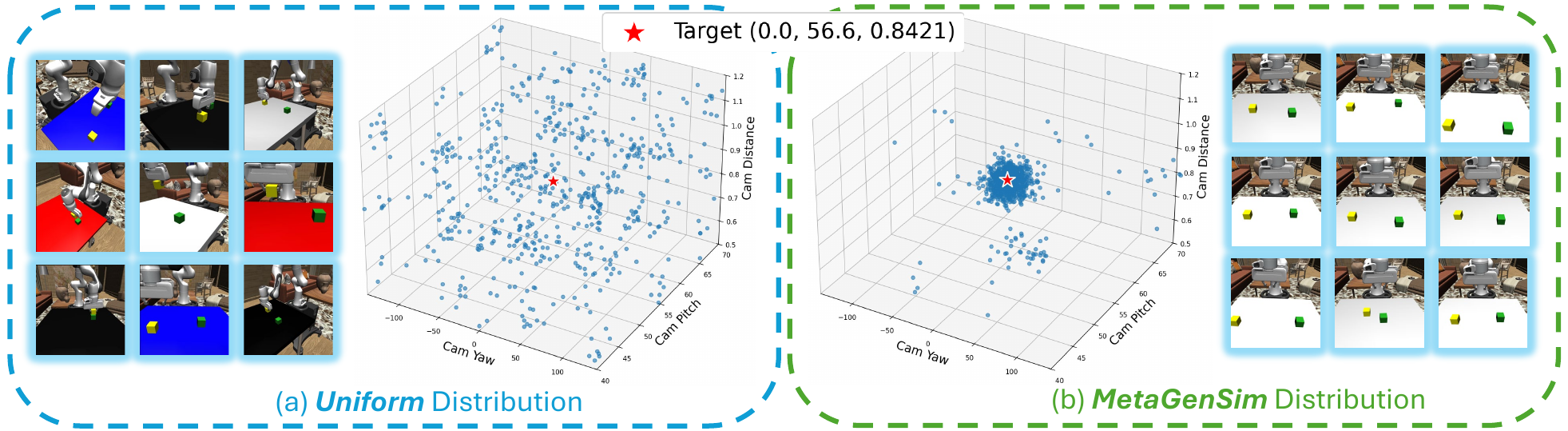}
  \caption{\ourmethod with 3D camera poses for the \textit{stackCube} task. Beyond the 1D yaw parameter used in the main real-world experiments, \ourmethod can optimize over yaw, pitch, and distance.}
  \label{fig:high-dim-cam}
\end{figure*}

\textbf{1D vs. 3D camera poses.}
The main real-world experiments use a 1D camera-yaw parameter because the physical setup fixes camera height and distance. To test whether the approach scales to higher-dimensional camera parameters, we extend the camera parameterization to 3D camera poses over yaw, pitch, and distance for \textit{stackCube}. As shown in Figure~\ref{fig:high-dim-cam}, \ourmethod still concentrates the generated data around target-aligned camera parameters, indicating that the density-estimation step can handle higher-dimensional camera-pose distributions.

\begin{figure*}[t]
  \centering
  \includegraphics[width=1.0\linewidth]{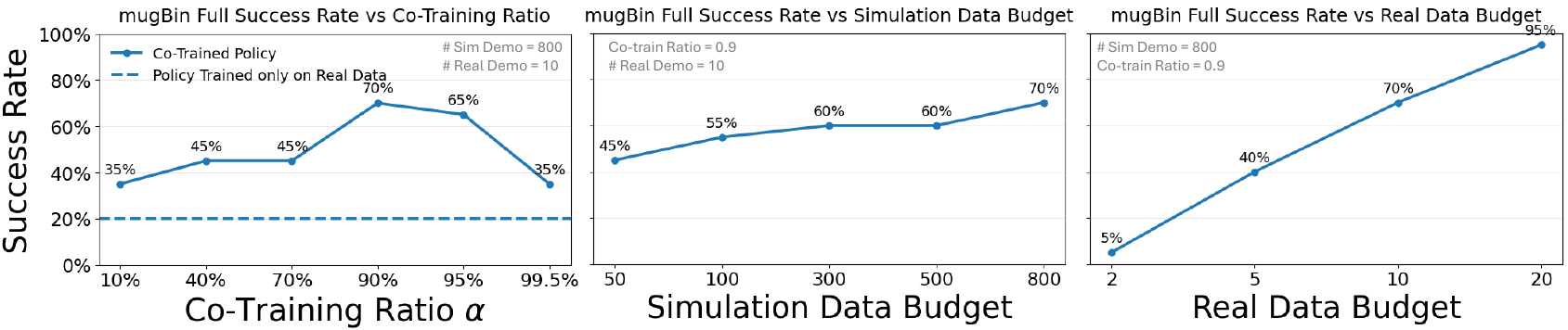}
  \caption{Sensitivity analysis for sim/real data budgets and the co-training ratio $\alpha$. Performance is best around $\alpha=0.9$; increasing the simulation budget helps only when sufficient real demonstrations are available.}
  \label{fig:rebuttal_ablation}
\end{figure*}

\textbf{Sensitivity analysis.}
We ablate the simulation budget, real-data budget, and co-training ratio $\alpha$ in Figure~\ref{fig:rebuttal_ablation}. The best performance occurs around $\alpha=0.9$, consistent with prior sim-and-real co-training findings and our main experimental setup. Increasing the simulation budget can improve performance, but the benefit remains limited by the amount of available real data, emphasizing that targeted simulation data complements rather than replaces real demonstrations.

\textbf{KDE bandwidth and temperature ablation.}

\begin{table}[t]
    \centering
    \scriptsize
    \caption{Ablations on hyperparameters of the data generation process~(\ref{sec:method-kde}) for the \textit{stackCube} task. \texttt{LowBW} denotes a bandwidth of $\frac{1}{4}$, \texttt{HighBW} denotes a bandwidth of $2\times$, and \texttt{HighTemp} denotes a temperature of $4\times$. Moderate temperature and sufficient bandwidth yield the best co-training performance, while overly high temperature hurts performance.}
    \begin{tabular}{lc|ccc}
        \toprule
        \multicolumn{2}{c|}{\textbf{Target Task}} & \texttt{LowBW} & \texttt{HighBW} & \texttt{HighTemp} \\
        \midrule
        \multirow{2}{*}{{stackCube}} & Pick & \rescount{12} & \rescount{17} & \rescount{2} \\
                                           & Pick + Place & \rescount{7}  & \rescount{14} & \rescount{0} \\
        \bottomrule
    \end{tabular}
    \label{tab:ablation_stackcube}
    \vspace{-2em}
\end{table}

We vary the hyperparameters introduced in~\ref{sec:method-kde} and generate co-training datasets for the {stackCube} task, and report the resulting co-training performance in Table~\ref{tab:ablation_stackcube}. We observe that setting the temperature too high yields a highly random dataset with an overly broad range of parameters, which degrades co-training performance. For the KDE bandwidth, an excessively low value concentrates samples near the probing data, reducing coverage of the parameter space and consequently hurting performance. We do not observe an obvious performance drop with higher bandwidth in this setting, although very large bandwidths produce increasingly diverse generated data.

%% file: algorithms/alg.tex
\begin{algorithm*}[tb]
\caption{Meta-learning the Rater Network ($R_\eta$)}
\label{alg:metagensim_rater}
\begin{algorithmic}[1]
\State \textbf{Inputs:} Simulation probing dataset $\mathcal{D}_{\text{probe}}$; Real-world dataset split into $\mathcal{D}_{R, \text{train}}$ and $\mathcal{D}_{R, \text{val}}$; Rater Network $R_\eta$ with meta-parameters $\eta$; Policy $\pi_\theta$ with parameters $\theta$; $K$ outer steps; $T$ inner steps; co-training ratio $\alpha$.

\State $\eta_0 \leftarrow \textsc{InitRater}()$ \Comment{Initialize Rater Network.}
\State $\theta_0 \leftarrow \textsc{InitPolicy}()$ \Comment{Initialize policy parameters.}
\For{$k = 0 \ldots K-1$} \Comment{Outer loop (Meta-Optimization).}

    \State $\theta_{k+1} = \textsc{UpdatePolicy}(\theta_k, \eta_k)$ \Comment{Unroll policy training for $T$ steps.}
    
    \State $\mathcal{B}_{\text{val}} \sim \mathcal{D}_{R, \text{val}}$ \Comment{Sample validation batch from held-out real data.}
    
    \State $L_{\text{val}} = \sum_{(o, a) \in \mathcal{B}_{\text{val}}} \ell_{BC}(\pi_{\theta_{k+1}}(o), a)$ \Comment{Compute validation loss on real data.}
    
    \State $\eta_{k+1} = \textsc{Optimizer}_\text{outer}(\eta_k, \nabla_\eta L_{\text{val}})$ \Comment{Update Rater using meta-gradient.}
\EndFor

\State \textbf{Return} $\eta_K$

\item[]
\Function{UpdatePolicy}{$\theta_0$, $\eta$}
  \For{$t = 0 \ldots T-1$}  \Comment{Inner loop (Policy Training).}
    \State $\mathcal{B}_{\text{sim}} \sim \mathcal{D}_{\text{probe}}$, \quad $\mathcal{B}_{\text{real}} \sim \mathcal{D}_{R, \text{train}}$ \Comment{Sample sim and real batches according to co-training ratio $\alpha$.}
    \State $\mathcal{B} \leftarrow \mathcal{B}_{\text{sim}} \cup \mathcal{B}_{\text{real}}$ \Comment{Combine into a single batch.}
    
    \State $s_i = R_\eta(o_i, a_i)$ for all $(o_i, a_i) \in \mathcal{B}$ \Comment{Compute raw scores.}
    \State $w_i = \frac{\exp(s_i)}{\sum_{j \in \mathcal{B}} \exp(s_j)}$ \Comment{Compute normalized weights for all samples.}
    
    \State $g_t = \sum_{(o_i, a_i) \in \mathcal{B}} w_i \cdot \nabla_\theta \ell_{BC}(\pi_{\theta_t}(o_i), a_i)$ \Comment{Weighted gradient.}
    \State $\theta_{t + 1} = \textsc{Optimizer}_\text{inner}(\theta_t, g_t)$ \Comment{Update policy parameters.}
  \EndFor
  \State \textbf{Return} $\theta_{T}$
\EndFunction
\end{algorithmic}
\end{algorithm*}

%% file: tables/sim2real_scaling_unif_gen.tex
\begin{wraptable}{r}{{0.4\linewidth}}
    \centering
    \scriptsize
    \setlength{\tabcolsep}{2pt}
    \vspace{-0.5cm}
    \caption{Scaling up \uniformgen. We generate 8000 uniform demos ($5\times$ more than \uniformgen used in the main experiments) for \textit{stackCube} and \textit{mugBin}, and report the resulting co-training performance. Increasing the amount of uniformly generated data can degrade co-training performance.}
    \begin{tabular}{lc|cc}
        \toprule
        \multicolumn{2}{c|}{\textbf{Target Task}} & 1600 demos & 8000 demos \\
        \midrule
        \multirow{2}{*}{stackCube} & Pick & \rescount{7} & \rescount{7} \\
                                  & Pick+Place & \rescount{2} & \rescount{3} \\
        \midrule
        \multirow{2}{*}{mugBin}    & Pick & \rescount{4} & \rescount{2} \\
                                  & Pick+Place & \rescount{3} & \rescount{1} \\
        \bottomrule
    \end{tabular}
    \label{tab:sim2real_scaling_unif_gen}
    \vspace{-0.5cm}
\end{wraptable}

%% file: main.bbl
\begin{thebibliography}{64}
\providecommand{\natexlab}[1]{#1}
\providecommand{\url}[1]{\texttt{#1}}
\expandafter\ifx\csname urlstyle\endcsname\relax
  \providecommand{\doi}[1]{doi: #1}\else
  \providecommand{\doi}{doi: \begingroup \urlstyle{rm}\Url}\fi

\bibitem[Ebert et~al.(2021)Ebert, Yang, Schmeckpeper, Bucher, Georgakis,
  Daniilidis, Finn, and Levine]{ebert2021bridge}
F.~Ebert, Y.~Yang, K.~Schmeckpeper, B.~Bucher, G.~Georgakis, K.~Daniilidis,
  C.~Finn, and S.~Levine.
\newblock Bridge data: Boosting generalization of robotic skills with
  cross-domain datasets.
\newblock \emph{arXiv preprint arXiv:2109.13396}, 2021.

\bibitem[Brohan et~al.(2022)Brohan, Brown, Carbajal, Chebotar, Dabis, Finn,
  Gopalakrishnan, Hausman, Herzog, Hsu, Ibarz, Ichter, Irpan, Jackson,
  Jesmonth, Joshi, Julian, Kalashnikov, Kuang, Leal, Lee, Levine, Lu, Malla,
  Manjunath, Mordatch, Nachum, Parada, Peralta, Perez, Pertsch, Quiambao, Rao,
  Ryoo, Salazar, Sanketi, Sayed, Singh, Sontakke, Stone, Tan, Tran, Vanhoucke,
  Vega, Vuong, Xia, Xiao, Xu, Xu, Yu, and Zitkovich]{rt12022arxiv}
A.~Brohan, N.~Brown, J.~Carbajal, Y.~Chebotar, J.~Dabis, C.~Finn,
  K.~Gopalakrishnan, K.~Hausman, A.~Herzog, J.~Hsu, J.~Ibarz, B.~Ichter,
  A.~Irpan, T.~Jackson, S.~Jesmonth, N.~Joshi, R.~Julian, D.~Kalashnikov,
  Y.~Kuang, I.~Leal, K.-H. Lee, S.~Levine, Y.~Lu, U.~Malla, D.~Manjunath,
  I.~Mordatch, O.~Nachum, C.~Parada, J.~Peralta, E.~Perez, K.~Pertsch,
  J.~Quiambao, K.~Rao, M.~Ryoo, G.~Salazar, P.~Sanketi, K.~Sayed, J.~Singh,
  S.~Sontakke, A.~Stone, C.~Tan, H.~Tran, V.~Vanhoucke, S.~Vega, Q.~Vuong,
  F.~Xia, T.~Xiao, P.~Xu, S.~Xu, T.~Yu, and B.~Zitkovich.
\newblock Rt-1: Robotics transformer for real-world control at scale.
\newblock In \emph{arXiv preprint arXiv:2212.06817}, 2022.

\bibitem[Black et~al.(2026)Black, Brown, Driess, Esmail, Equi, Finn, Fusai,
  Groom, Hausman, Ichter, Jakubczak, Jones, Ke, Levine, Li-Bell, Mothukuri,
  Nair, Pertsch, Shi, Tanner, Vuong, Walling, Wang, and
  Zhilinsky]{black2026pi0visionlanguageactionflowmodel}
K.~Black, N.~Brown, D.~Driess, A.~Esmail, M.~Equi, C.~Finn, N.~Fusai, L.~Groom,
  K.~Hausman, B.~Ichter, S.~Jakubczak, T.~Jones, L.~Ke, S.~Levine, A.~Li-Bell,
  M.~Mothukuri, S.~Nair, K.~Pertsch, L.~X. Shi, J.~Tanner, Q.~Vuong,
  A.~Walling, H.~Wang, and U.~Zhilinsky.
\newblock $\pi_0$: A vision-language-action flow model for general robot
  control, 2026.
\newblock URL \url{https://arxiv.org/abs/2410.24164}.

\bibitem[Dei(2022)]{Deitke2022ProcTHORLa}
{ProcTHOR: Large-Scale Embodied AI Using Procedural Generation}.
\newblock \emph{NeurIPS 2022}, 2022.

\bibitem[Liang et~al.(2024)Liang, Wang, Wang, Bastani, Jayaraman, and
  Ma]{liang2024eurekaverse}
W.~Liang, S.~Wang, H.-J. Wang, O.~Bastani, D.~Jayaraman, and Y.~J. Ma.
\newblock Eurekaverse: Environment curriculum generation via large language
  models.
\newblock \emph{arXiv preprint arXiv:2411.01775}, 2024.

\bibitem[Deshpande et~al.(2025)Deshpande, Deng, Ray, Salvador, Han, Duan, Zeng,
  Zhu, Krishna, and Hendrix]{deshpande2025graspmolmo}
A.~Deshpande, Y.~Deng, A.~Ray, J.~Salvador, W.~Han, J.~Duan, K.-H. Zeng,
  Y.~Zhu, R.~Krishna, and R.~Hendrix.
\newblock Graspmolmo: Generalizable task-oriented grasping via large-scale
  synthetic data generation, 2025.
\newblock URL \url{https://arxiv.org/abs/2505.13441}.

\bibitem[Katara et~al.(2024)Katara, Xian, and Fragkiadaki]{katara2024gen2sim}
P.~Katara, Z.~Xian, and K.~Fragkiadaki.
\newblock Gen2sim: Scaling up robot learning in simulation with generative
  models.
\newblock In \emph{2024 IEEE International Conference on Robotics and
  Automation (ICRA)}, pages 6672--6679. IEEE, 2024.

\bibitem[Nasiriany et~al.(2024)Nasiriany, Maddukuri, Zhang, Parikh, Lo, Joshi,
  Mandlekar, and Zhu]{robocasa2024}
S.~Nasiriany, A.~Maddukuri, L.~Zhang, A.~Parikh, A.~Lo, A.~Joshi, A.~Mandlekar,
  and Y.~Zhu.
\newblock Robocasa: Large-scale simulation of everyday tasks for generalist
  robots.
\newblock In \emph{Robotics: Science and Systems (RSS)}, 2024.

\bibitem[Mandlekar et~al.(2023)Mandlekar, Nasiriany, Wen, Akinola, Narang, Fan,
  Zhu, and Fox]{mandlekar2023mimicgen}
A.~Mandlekar, S.~Nasiriany, B.~Wen, I.~Akinola, Y.~Narang, L.~Fan, Y.~Zhu, and
  D.~Fox.
\newblock Mimicgen: A data generation system for scalable robot learning using
  human demonstrations.
\newblock \emph{arXiv preprint arXiv:2310.17596}, 2023.

\bibitem[Jiang et~al.(2024)Jiang, Xie, Lin, Xu, Wan, Mandlekar, Fan, and
  Zhu]{jiang2024dexmimicgen}
Z.~Jiang, Y.~Xie, K.~Lin, Z.~Xu, W.~Wan, A.~Mandlekar, L.~Fan, and Y.~Zhu.
\newblock Dexmimicgen: Automated data generation for bimanual dexterous
  manipulation via imitation learning.
\newblock \emph{arXiv preprint arXiv:2410.24185}, 2024.

\bibitem[Garrett et~al.(2024)Garrett, Mandlekar, Wen, and
  Fox]{garrett2024skillgen}
C.~R. Garrett, A.~Mandlekar, B.~Wen, and D.~Fox.
\newblock Skillgen: Automated demonstration generation for efficient skill
  learning and deployment.
\newblock In \emph{8th Annual Conference on Robot Learning}, 2024.
\newblock URL \url{https://openreview.net/forum?id=YOFrRTDC6d}.

\bibitem[Cheng et~al.(2023)Cheng, Garrett, Mandlekar, and Xu]{cheng2023nod}
S.~Cheng, C.~Garrett, A.~Mandlekar, and D.~Xu.
\newblock Nod-tamp: Multi-step manipulation planning with neural object
  descriptors.
\newblock \emph{arXiv preprint arXiv:2311.01530}, 2023.

\bibitem[Lin et~al.(2026)Lin, Mandlekar, Garrett, Chernyadev, Fang, Ding, Xie,
  Tran, Fan, and Zhu]{lin2026humanoidmimicgendatagenerationlocomanipulation}
K.~Lin, A.~Mandlekar, C.~R. Garrett, N.~Chernyadev, Y.~Fang, R.~Ding, Y.~Xie,
  J.~Tran, L.~Fan, and Y.~Zhu.
\newblock Humanoidmimicgen: Data generation for loco-manipulation via
  whole-body planning, 2026.
\newblock URL \url{https://arxiv.org/abs/2605.27724}.

\bibitem[Jain et~al.(2025)Jain, Zhang, Arora, Chen, Torne, Irshad, Zakharov,
  Wang, Levine, Finn, Ma, Shah, Gupta, and
  Pertsch]{jain2025polarisscalablerealtosimevaluations}
A.~Jain, M.~Zhang, K.~Arora, W.~Chen, M.~Torne, M.~Z. Irshad, S.~Zakharov,
  Y.~Wang, S.~Levine, C.~Finn, W.-C. Ma, D.~Shah, A.~Gupta, and K.~Pertsch.
\newblock Polaris: Scalable real-to-sim evaluations for generalist robot
  policies, 2025.
\newblock URL \url{https://arxiv.org/abs/2512.16881}.

\bibitem[Ankile et~al.(2024)Ankile, Simeonov, Shenfeld, Torne, and
  Agrawal]{ankile2024imitationrefinementresidual}
L.~Ankile, A.~Simeonov, I.~Shenfeld, M.~Torne, and P.~Agrawal.
\newblock From imitation to refinement -- residual rl for precise assembly,
  2024.
\newblock URL \url{https://arxiv.org/abs/2407.16677}.

\bibitem[Wei et~al.(2025)Wei, Agarwal, Chen, Bosworth, Pfaff, and
  Tedrake]{wei2025simandrealcotraining}
A.~Wei, A.~Agarwal, B.~Chen, R.~Bosworth, N.~Pfaff, and R.~Tedrake.
\newblock Empirical analysis of sim-and-real cotraining of diffusion policies
  for planar pushing from pixels.
\newblock 2025.
\newblock URL \url{https://arxiv.org/abs/2503.22634}.

\bibitem[Maddukuri et~al.(2025)Maddukuri, Jiang, Chen, Nasiriany, Xie, Fang,
  Huang, Wang, Xu, Chernyadev, et~al.]{maddukuri2025simandreal}
A.~Maddukuri, Z.~Jiang, L.~Y. Chen, S.~Nasiriany, Y.~Xie, Y.~Fang, W.~Huang,
  Z.~Wang, Z.~Xu, N.~Chernyadev, et~al.
\newblock Sim-and-real co-training: A simple recipe for vision-based robotic
  manipulation.
\newblock \emph{arXiv preprint arXiv:2503.24361}, 2025.

\bibitem[Cheng et~al.(2025)Cheng, Ma, Chen, Mandlekar, Garrett, and
  Xu]{cheng2025generalizable}
S.~Cheng, L.~Ma, Z.~Chen, A.~Mandlekar, C.~R. Garrett, and D.~Xu.
\newblock Generalizable domain adaptation for sim-and-real policy co-training.
\newblock In \emph{The Thirty-ninth Annual Conference on Neural Information
  Processing Systems}, 2025.

\bibitem[Lei et~al.(2026)Lei, Liu, Maddukuri, Jiang, and
  Zhu]{lei2026mechanisticanalysis}
Y.~Lei, M.~Liu, A.~Maddukuri, Z.~Jiang, and Y.~Zhu.
\newblock A mechanistic analysis of sim-and-real co-training in generative
  robot policies, 2026.
\newblock URL \url{https://arxiv.org/abs/2604.13645}.

\bibitem[Anonymous(2025)]{anonymous2025robocasa}
Anonymous.
\newblock Robocasa365: A large-scale simulation framework for training and
  benchmarking generalist robots.
\newblock In \emph{Submitted to The Fourteenth International Conference on
  Learning Representations}, 2025.
\newblock URL \url{https://openreview.net/forum?id=tQJYKwc3n4}.
\newblock under review.

\bibitem[Mu et~al.(2024)Mu, Guo, Xu, Goyal, Su, Fox, and Garg]{mu2024adademo}
T.~Mu, Y.~Guo, J.~Xu, A.~Goyal, H.~Su, D.~Fox, and A.~Garg.
\newblock Adademo: Data-efficient demonstration expansion for generalist
  robotic agent.
\newblock \emph{arXiv preprint arXiv:2404.07428}, 2024.

\bibitem[Gao et~al.(2024)Gao, Xie, Xiao, Finn, and Sadigh]{gao2024efficient}
J.~Gao, A.~Xie, T.~Xiao, C.~Finn, and D.~Sadigh.
\newblock Efficient data collection for robotic manipulation via compositional
  generalization.
\newblock \emph{arXiv preprint arXiv:2403.05110}, 2024.

\bibitem[Torne et~al.(2024)Torne, Simeonov, Li, Chan, Chen, Gupta, and
  Agrawal]{torne2024reconciling}
M.~Torne, A.~Simeonov, Z.~Li, A.~Chan, T.~Chen, A.~Gupta, and P.~Agrawal.
\newblock Reconciling reality through simulation: A real-to-sim-to-real
  approach for robust manipulation.
\newblock \emph{arXiv preprint arXiv:2403.03949}, 2024.

\bibitem[Tobin et~al.(2017)Tobin, Fong, Ray, Schneider, Zaremba, and
  Abbeel]{tobin2017domain}
J.~Tobin, R.~Fong, A.~Ray, J.~Schneider, W.~Zaremba, and P.~Abbeel.
\newblock Domain randomization for transferring deep neural networks from
  simulation to the real world.
\newblock In \emph{2017 IEEE/RSJ international conference on intelligent robots
  and systems (IROS)}, pages 23--30. IEEE, 2017.

\bibitem[Dass et~al.(2025)Dass, Khaddaj, Engstrom, Madry, Ilyas, and
  Mart{\'\i}n-Mart{\'\i}n]{dass2025datamil}
S.~Dass, A.~Khaddaj, L.~Engstrom, A.~Madry, A.~Ilyas, and
  R.~Mart{\'\i}n-Mart{\'\i}n.
\newblock Datamil: Selecting data for robot imitation learning with datamodels.
\newblock \emph{arXiv preprint arXiv:2505.09603}, 2025.

\bibitem[Agia et~al.(2025)Agia, Sinha, Yang, Antonova, Pavone, Nishimura,
  Itkina, and Bohg]{agia2025cupid}
C.~Agia, R.~Sinha, J.~Yang, R.~Antonova, M.~Pavone, H.~Nishimura, M.~Itkina,
  and J.~Bohg.
\newblock Cupid: Curating data your robot loves with influence functions.
\newblock \emph{arXiv preprint arXiv:2506.19121}, 2025.

\bibitem[Kuhar et~al.(2023)Kuhar, Cheng, Chopra, Bronars, and
  Xu]{kuhar2023learning}
S.~Kuhar, S.~Cheng, S.~Chopra, M.~Bronars, and D.~Xu.
\newblock Learning to discern: Imitating heterogeneous human demonstrations
  with preference and representation learning.
\newblock In \emph{Conference on Robot Learning}, pages 1437--1449. PMLR, 2023.

\bibitem[Chen et~al.(2025)Chen, Lessing, Liu, and Finn]{chen2025curating}
A.~S. Chen, A.~M. Lessing, Y.~Liu, and C.~Finn.
\newblock Curating demonstrations using online experience.
\newblock \emph{arXiv preprint arXiv:2503.03707}, 2025.

\bibitem[Memmel et~al.(2024)Memmel, Berg, Chen, Gupta, and
  Francis]{memmel2024strap}
M.~Memmel, J.~Berg, B.~Chen, A.~Gupta, and J.~Francis.
\newblock Strap: Robot sub-trajectory retrieval for augmented policy learning.
\newblock \emph{arXiv preprint arXiv:2412.15182}, 2024.

\bibitem[NVIDIA et~al.(2024)NVIDIA, Bala, Cui, Ding, Ge, Hao, Hasselgren,
  Huffman, Jin, Lewis, Li, Lin, Lin, Lin, Liu, Luo, Ma, Munkberg, Shi, Wei,
  Xiang, Xu, Zeng, and Zhang]{nvidia2024edify3d}
NVIDIA, M.~Bala, Y.~Cui, Y.~Ding, Y.~Ge, Z.~Hao, J.~Hasselgren, J.~Huffman,
  J.~Jin, J.~Lewis, Z.~Li, C.-H. Lin, Y.-C. Lin, T.-Y. Lin, M.-Y. Liu, A.~Luo,
  Q.~Ma, J.~Munkberg, S.~Shi, F.~Wei, D.~Xiang, J.~Xu, X.~Zeng, and Q.~Zhang.
\newblock Edify 3d: Scalable high-quality 3d asset generation.
\newblock \emph{arXiv preprint arXiv:2411.07135}, 2024.

\bibitem[Chen et~al.(2023)Chen, Chen, Jiao, and Jia]{chen2023fantasia3d}
R.~Chen, Y.~Chen, N.~Jiao, and K.~Jia.
\newblock Fantasia3d: Disentangling geometry and appearance for high-quality
  text-to-3d content creation.
\newblock In \emph{Proceedings of the IEEE/CVF International Conference on
  Computer Vision (ICCV)}, pages 22246--22256, October 2023.

\bibitem[Cheng et~al.(2024)Cheng, Garrett, Mandlekar, and Xu]{cheng2024nodtamp}
S.~Cheng, C.~R. Garrett, A.~Mandlekar, and D.~Xu.
\newblock {NOD}-{TAMP}: Generalizable long-horizon planning with neural object
  descriptors.
\newblock In \emph{8th Annual Conference on Robot Learning}, 2024.
\newblock URL \url{https://openreview.net/forum?id=rThtgkXuvZ}.

\bibitem[Xue et~al.(2025)Xue, Deng, Chen, Wang, Yuan, and Xu]{xue2025demogen}
Z.~Xue, S.~Deng, Z.~Chen, Y.~Wang, Z.~Yuan, and H.~Xu.
\newblock Demogen: Synthetic demonstration generation for data-efficient
  visuomotor policy learning.
\newblock \emph{arXiv preprint arXiv:2502.16932}, 2025.

\bibitem[Wang et~al.(2023)Wang, Xian, Chen, Wang, Wang, Fragkiadaki, Erickson,
  Held, and Gan]{wang2023robogen}
Y.~Wang, Z.~Xian, F.~Chen, T.-H. Wang, Y.~Wang, K.~Fragkiadaki, Z.~Erickson,
  D.~Held, and C.~Gan.
\newblock Robogen: Towards unleashing infinite data for automated robot
  learning via generative simulation.
\newblock \emph{arXiv preprint arXiv:2311.01455}, 2023.

\bibitem[Hua et~al.()Hua, Liu, Macaluso, Lin, Zhang, Xu, and Wang]{huagensim2}
P.~Hua, M.~Liu, A.~Macaluso, Y.~Lin, W.~Zhang, H.~Xu, and L.~Wang.
\newblock Gensim2: Scaling robot data generation with multi-modal and reasoning
  llms.
\newblock In \emph{8th Annual Conference on Robot Learning}.

\bibitem[Dai et~al.(2024)Dai, Wong, Jiang, Wang, Gokmen, Zhang, Wu, and
  Fei-Fei]{dai2024automated}
T.~Dai, J.~Wong, Y.~Jiang, C.~Wang, C.~Gokmen, R.~Zhang, J.~Wu, and L.~Fei-Fei.
\newblock Automated creation of digital cousins for robust policy learning.
\newblock \emph{arXiv preprint arXiv:2410.07408}, 2024.

\bibitem[Tian et~al.(2025)Tian, Yang, Xie, Cai, Shi, Gao, Liu, Jiang, Qiu,
  Yuan, Li, Wang, Cai, Zeng, Dong, and
  Pang]{tian2025interndataa1pioneeringhighfidelitysynthetic}
Y.~Tian, Y.~Yang, Y.~Xie, Z.~Cai, X.~Shi, N.~Gao, H.~Liu, X.~Jiang, Z.~Qiu,
  F.~Yuan, Y.~Li, P.~Wang, J.~Cai, J.~Zeng, H.~Dong, and J.~Pang.
\newblock Interndata-a1: Pioneering high-fidelity synthetic data for
  pre-training generalist policy, 2025.
\newblock URL \url{https://arxiv.org/abs/2511.16651}.

\bibitem[Li et~al.(2025)Li, Xu, Bahety, Yin, Jiang, Huang, Wong, Garlanka,
  Gokmen, Zhang, et~al.]{li2025momagen}
C.~Li, M.~Xu, A.~Bahety, H.~Yin, Y.~Jiang, H.~Huang, J.~Wong, S.~Garlanka,
  C.~Gokmen, R.~Zhang, et~al.
\newblock Momagen: Generating demonstrations under soft and hard constraints
  for multi-step bimanual mobile manipulation.
\newblock \emph{arXiv preprint arXiv:2510.18316}, 2025.

\bibitem[Gong et~al.(2025)Gong, Zhang, Shang, Minniti, Patel, Pepe, Yan,
  Gundogdu, Kapelyukh, Abbas, et~al.]{gong2025anytask}
R.~Gong, X.~Zhang, J.~Shang, M.~V. Minniti, J.~Patel, V.~Pepe, R.~Yan,
  A.~Gundogdu, I.~Kapelyukh, A.~Abbas, et~al.
\newblock Anytask: an automated task and data generation framework for
  advancing sim-to-real policy learning.
\newblock \emph{arXiv preprint arXiv:2512.17853}, 2025.

\bibitem[Saxena et~al.(2024)Saxena, koga, and Xu]{saxena2024cdm}
V.~Saxena, Y.~koga, and D.~Xu.
\newblock C3{DM}: Constrained-context conditional diffusion models for
  imitation learning.
\newblock \emph{Transactions on Machine Learning Research}, 2024.
\newblock ISSN 2835-8856.
\newblock URL \url{https://openreview.net/forum?id=jcleXdnRA1}.

\bibitem[Peng et~al.(2018)Peng, Andrychowicz, Zaremba, and Abbeel]{peng2018sim}
X.~B. Peng, M.~Andrychowicz, W.~Zaremba, and P.~Abbeel.
\newblock Sim-to-real transfer of robotic control with dynamics randomization.
\newblock In \emph{2018 IEEE international conference on robotics and
  automation (ICRA)}, pages 3803--3810. IEEE, 2018.

\bibitem[Chebotar et~al.(2019)Chebotar, Handa, Makoviychuk, Macklin, Issac,
  Ratliff, and Fox]{chebotar2019closing}
Y.~Chebotar, A.~Handa, V.~Makoviychuk, M.~Macklin, J.~Issac, N.~Ratliff, and
  D.~Fox.
\newblock Closing the sim-to-real loop: Adapting simulation randomization with
  real world experience.
\newblock In \emph{2019 International Conference on Robotics and Automation
  (ICRA)}, pages 8973--8979. IEEE, 2019.

\bibitem[Du et~al.(2021)Du, Watkins, Darrell, Abbeel, and Pathak]{9562091}
Y.~Du, O.~Watkins, T.~Darrell, P.~Abbeel, and D.~Pathak.
\newblock Auto-tuned sim-to-real transfer.
\newblock In \emph{2021 IEEE International Conference on Robotics and
  Automation (ICRA)}, pages 1290--1296, 2021.
\newblock \doi{10.1109/ICRA48506.2021.9562091}.

\bibitem[Ren et~al.(2023)Ren, Dai, Burchfiel, and Majumdar]{ren2023adaptsim}
A.~Z. Ren, H.~Dai, B.~Burchfiel, and A.~Majumdar.
\newblock Adaptsim: Task-driven simulation adaptation for sim-to-real transfer.
\newblock \emph{arXiv preprint arXiv:2302.04903}, 2023.

\bibitem[Memmel et~al.(2024)Memmel, Wagenmaker, Zhu, Yin, Fox, and
  Gupta]{memmel2024asid}
M.~Memmel, A.~Wagenmaker, C.~Zhu, P.~Yin, D.~Fox, and A.~Gupta.
\newblock Asid: Active exploration for system identification in robotic
  manipulation.
\newblock \emph{arXiv preprint arXiv:2404.12308}, 2024.

\bibitem[Ramos et~al.(2019)Ramos, Possas, and Fox]{ramos2019bayessim}
F.~Ramos, R.~C. Possas, and D.~Fox.
\newblock Bayessim: adaptive domain randomization via probabilistic inference
  for robotics simulators.
\newblock \emph{arXiv preprint arXiv:1906.01728}, 2019.

\bibitem[Lin et~al.(2024)Lin, Cui, Xie, Hua, and Sadigh]{lin2024flowretrieval}
L.-H. Lin, Y.~Cui, A.~Xie, T.~Hua, and D.~Sadigh.
\newblock Flowretrieval: Flow-guided data retrieval for few-shot imitation
  learning.
\newblock \emph{arXiv preprint arXiv:2408.16944}, 2024.

\bibitem[Hejna et~al.(2025)Hejna, Mirchandani, Balakrishna, Xie, Wahid,
  Tompson, Sanketi, Shah, Devin, and Sadigh]{hejna2025robot}
J.~Hejna, S.~Mirchandani, A.~Balakrishna, A.~Xie, A.~Wahid, J.~Tompson,
  P.~Sanketi, D.~Shah, C.~Devin, and D.~Sadigh.
\newblock Robot data curation with mutual information estimators.
\newblock \emph{arXiv preprint arXiv:2502.08623}, 2025.

\bibitem[Xie et~al.(2025)Xie, Chand, Sadigh, and Hejna]{xie2025data}
A.~Xie, R.~Chand, D.~Sadigh, and J.~Hejna.
\newblock Data retrieval with importance weights for few-shot imitation
  learning.
\newblock In \emph{9th Annual Conference on Robot Learning}, 2025.
\newblock URL \url{https://arxiv.org/abs/2509.01657}.

\bibitem[Du et~al.(2023)Du, Nair, Sadigh, and Finn]{du2023behavior}
M.~Du, S.~Nair, D.~Sadigh, and C.~Finn.
\newblock Behavior retrieval: Few-shot imitation learning by querying unlabeled
  datasets.
\newblock \emph{arXiv preprint arXiv:2304.08742}, 2023.

\bibitem[Engstrom et~al.(2025)Engstrom, Ilyas, Chen, Feldmann, Moses, and
  Madry]{engstrom2025optimizing}
L.~Engstrom, A.~Ilyas, B.~Chen, A.~Feldmann, W.~Moses, and A.~Madry.
\newblock Optimizing ml training with metagradient descent.
\newblock \emph{arXiv preprint arXiv:2503.13751}, 2025.

\bibitem[Calian et~al.(2025)Calian, Farquhar, Kemaev, Zintgraf, Hessel, Shar,
  Oh, Gy{\"o}rgy, Schaul, Dean, et~al.]{calian2025datarater}
D.~A. Calian, G.~Farquhar, I.~Kemaev, L.~M. Zintgraf, M.~Hessel, J.~Shar,
  J.~Oh, A.~Gy{\"o}rgy, T.~Schaul, J.~Dean, et~al.
\newblock Datarater: Meta-learned dataset curation.
\newblock \emph{arXiv preprint arXiv:2505.17895}, 2025.

\bibitem[Scott(2015)]{scott2015multivariate}
D.~W. Scott.
\newblock \emph{Multivariate density estimation: theory, practice, and
  visualization}.
\newblock John Wiley \& Sons, 2015.

\bibitem[Saxena et~al.(2025)Saxena, Bronars, Arachchige, Wang, Shin, Nasiriany,
  Mandlekar, and Xu]{saxena2025mimiclabs}
V.~Saxena, M.~Bronars, N.~R. Arachchige, K.~Wang, W.~C. Shin, S.~Nasiriany,
  A.~Mandlekar, and D.~Xu.
\newblock What matters in learning from large-scale datasets for robot
  manipulation.
\newblock In \emph{The Thirteenth International Conference on Learning
  Representations}, 2025.
\newblock URL \url{https://arxiv.org/pdf/2506.13536}.

\bibitem[Chi et~al.(2023)Chi, Feng, Du, Xu, Cousineau, Burchfiel, and
  Song]{chi2023diffusionpolicy}
C.~Chi, S.~Feng, Y.~Du, Z.~Xu, E.~Cousineau, B.~Burchfiel, and S.~Song.
\newblock Diffusion policy: Visuomotor policy learning via action diffusion.
\newblock In \emph{Proceedings of Robotics: Science and Systems (RSS)}, 2023.

\bibitem[Chi et~al.(2024)Chi, Xu, Feng, Cousineau, Du, Burchfiel, Tedrake, and
  Song]{chi2024diffusionpolicy}
C.~Chi, Z.~Xu, S.~Feng, E.~Cousineau, Y.~Du, B.~Burchfiel, R.~Tedrake, and
  S.~Song.
\newblock Diffusion policy: Visuomotor policy learning via action diffusion.
\newblock \emph{The International Journal of Robotics Research}, 2024.

\bibitem[Intelligence et~al.(2025)Intelligence, Black, Brown, Darpinian,
  Dhabalia, Driess, Esmail, Equi, Finn, Fusai, Galliker, Ghosh, Groom, Hausman,
  Ichter, Jakubczak, Jones, Ke, LeBlanc, Levine, Li-Bell, Mothukuri, Nair,
  Pertsch, Ren, Shi, Smith, Springenberg, Stachowicz, Tanner, Vuong, Walke,
  Walling, Wang, Yu, and
  Zhilinsky]{intelligence2025pi05visionlanguageactionmodelopenworld}
P.~Intelligence, K.~Black, N.~Brown, J.~Darpinian, K.~Dhabalia, D.~Driess,
  A.~Esmail, M.~Equi, C.~Finn, N.~Fusai, M.~Y. Galliker, D.~Ghosh, L.~Groom,
  K.~Hausman, B.~Ichter, S.~Jakubczak, T.~Jones, L.~Ke, D.~LeBlanc, S.~Levine,
  A.~Li-Bell, M.~Mothukuri, S.~Nair, K.~Pertsch, A.~Z. Ren, L.~X. Shi,
  L.~Smith, J.~T. Springenberg, K.~Stachowicz, J.~Tanner, Q.~Vuong, H.~Walke,
  A.~Walling, H.~Wang, L.~Yu, and U.~Zhilinsky.
\newblock $\pi_{0.5}$: a vision-language-action model with open-world
  generalization, 2025.
\newblock URL \url{https://arxiv.org/abs/2504.16054}.

\bibitem[Kingma and Ba(2017)]{kingma2017adammethodstochasticoptimization}
D.~P. Kingma and J.~Ba.
\newblock Adam: A method for stochastic optimization, 2017.
\newblock URL \url{https://arxiv.org/abs/1412.6980}.

\bibitem[Ren* et~al.(2023)Ren*, Feng*, Liu*, Pan*, Fu, Mai, and
  Yang]{JMLR:TorchOpt}
J.~Ren*, X.~Feng*, B.~Liu*, X.~Pan*, Y.~Fu, L.~Mai, and Y.~Yang.
\newblock Torchopt: An efficient library for differentiable optimization.
\newblock \emph{Journal of Machine Learning Research}, 24\penalty0
  (367):\penalty0 1--14, 2023.
\newblock URL \url{http://jmlr.org/papers/v24/23-0191.html}.

\bibitem[Mandlekar et~al.(2021)Mandlekar, Xu, Wong, Nasiriany, Wang, Kulkarni,
  Fei-Fei, Savarese, Zhu, and Mart{\'\i}n-Mart{\'\i}n]{mandlekar2021matters}
A.~Mandlekar, D.~Xu, J.~Wong, S.~Nasiriany, C.~Wang, R.~Kulkarni, L.~Fei-Fei,
  S.~Savarese, Y.~Zhu, and R.~Mart{\'\i}n-Mart{\'\i}n.
\newblock What matters in learning from offline human demonstrations for robot
  manipulation.
\newblock \emph{arXiv preprint arXiv:2108.03298}, 2021.

\bibitem[Levine et~al.(2016)Levine, Finn, Darrell, and Abbeel]{levine2016end}
S.~Levine, C.~Finn, T.~Darrell, and P.~Abbeel.
\newblock End-to-end training of deep visuomotor policies.
\newblock \emph{Journal of Machine Learning Research}, 17\penalty0
  (39):\penalty0 1--40, 2016.

\bibitem[Oquab et~al.(2023)Oquab, Darcet, Moutakanni, Vo, Szafraniec, Khalidov,
  Fernandez, Haziza, Massa, El-Nouby, et~al.]{oquab2023dinov2}
M.~Oquab, T.~Darcet, T.~Moutakanni, H.~Vo, M.~Szafraniec, V.~Khalidov,
  P.~Fernandez, D.~Haziza, F.~Massa, A.~El-Nouby, et~al.
\newblock Dinov2: Learning robust visual features without supervision.
\newblock \emph{arXiv preprint arXiv:2304.07193}, 2023.

\bibitem[Maaten and Hinton(2008)]{maaten2008visualizing}
L.~v.~d. Maaten and G.~Hinton.
\newblock Visualizing data using t-sne.
\newblock \emph{Journal of machine learning research}, 9\penalty0
  (Nov):\penalty0 2579--2605, 2008.

\bibitem[Chi et~al.(2022)Chi, Burchfiel, Cousineau, Feng, and
  Song]{Chi2022IterativeRP}
C.~Chi, B.~Burchfiel, E.~A. Cousineau, S.~Feng, and S.~Song.
\newblock Iterative residual policy: For goal-conditioned dynamic manipulation
  of deformable objects.
\newblock \emph{The International Journal of Robotics Research}, 43:\penalty0
  389 -- 404, 2022.

\end{thebibliography}
